\documentclass[10pt,twocolumn,letterpaper]{article}

\usepackage{wacv}              

\usepackage{graphicx}
\usepackage{amsmath}
\usepackage{amssymb}
\usepackage{booktabs}
\usepackage{multirow}
\usepackage[export]{adjustbox}
\usepackage{tabularx} 
\usepackage{caption}
\usepackage{makecell}
\usepackage[accsupp]{axessibility}  

\usepackage[pagebackref,breaklinks,colorlinks]{hyperref}

\usepackage[capitalize]{cleveref}
\crefname{section}{Sec.}{Secs.}
\Crefname{section}{Section}{Sections}
\Crefname{table}{Table}{Tables}
\crefname{table}{Tab.}{Tabs.}

\def\wacvPaperID{1701} 
\def\confName{WACV}
\def\confYear{2024}

\begin{document}

\title{Global Occlusion-Aware Transformer for Robust Stereo Matching}

\author{Zihua Liu$^{1}$, Yizhou Li$^{2}$, and Masatoshi Okutomi$^{3}$ \\
Tokyo Institute of Technology, Japan \\
{\tt\small \{zliu$^{1}$,yli$^{2}$\}@ok.sc.e.titech.ac.jp,mxo@ctrl.titech.ac.jp$^{3}$}
}

\maketitle

\begin{abstract}
Despite the remarkable progress facilitated by learning-based stereo-matching algorithms, the performance in the ill-conditioned regions, such as the occluded regions, remains a bottleneck. To address this issue, this paper introduces a novel attention-based stereo-matching network called \underline{G}lobal \underline{O}cclusion-\underline{A}ware \underline{T}ransformer (\textit{GOAT}) to exploit long-range dependency and occlusion-awareness global context for disparity estimation. In the \textit{GOAT} architecture, a parallel disparity and occlusion estimation module~(PDO) is proposed to estimate the initial disparity map and the occlusion mask using a parallel attention mechanism. To further enhance the disparity estimates in the occluded regions, an occlusion-aware global aggregation module (OGA) is proposed. This module aims to refine the disparity in the occluded regions by leveraging restricted global correlation within the focus scope of the occluded areas. Extensive experiments were conducted on several public benchmark datasets including SceneFlow \cite{dispnetc}, KITTI 2015 \cite{Kitti2015}, and Middlebury \cite{middlebury2014}. The results show that proposed \textit{GOAT} demonstrates outstanding performance among all benchmarks, particularly in the occluded regions. Code is available at \href{https://github.com/Magicboomliu/GOAT}{https://github.com/Magicboomliu/GOAT}.
\end{abstract}
\begin{figure}
	\centering
\includegraphics[width=1.00\linewidth]{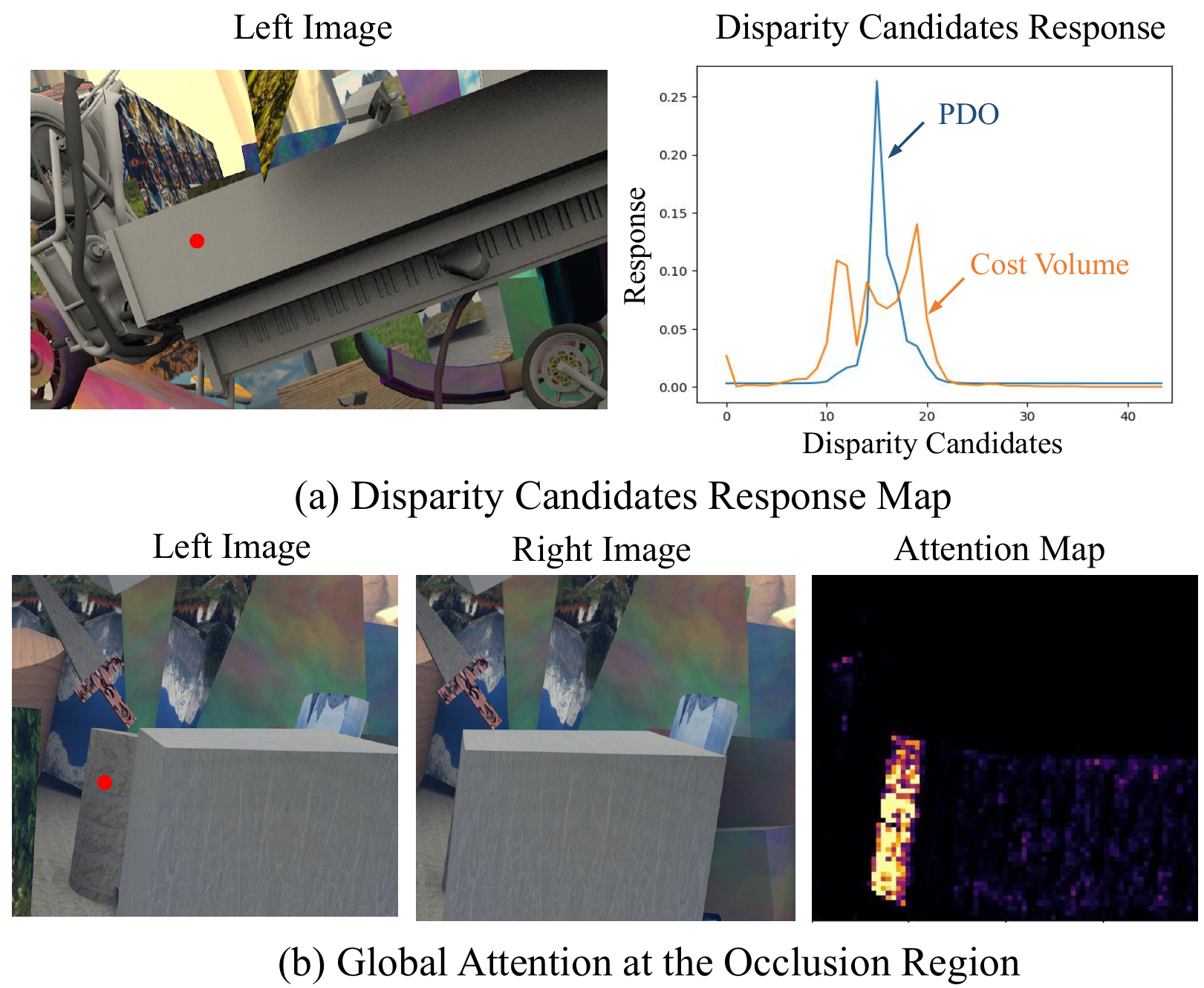}
	\caption{\textbf{(a)} Visualization of estimated response for disparity candidates using proposed \textit{PDO}. Compared with a cost volume method (\textcolor[rgb]{1,0.4,0}{orange}), the \textit{PDO} 
(\textcolor[rgb]{0.0,0.0,1.0}{blue}) can alleviate matching ambiguity in texture-less regions and show a single peak waveform. \textbf{(b)} Visualization of global attention map in the occluded regions using the proposed \textit{OGA}. }
	\label{fig:opening}
	\vspace{-0.8 em}
\end{figure}

\section{Introduction}
Stereo-matching is one of the most fundamental tasks in computer vision. It is to infer depth from a given pair of stereo images taken by a binocular camera, which is closely related to applications like robotic navigation~\cite{stereorobotic}, autonomous driving~\cite{stereodriving}, augmented reality~\cite{stereoar}, and so on.

Recently, the rapid development of convolutional neural networks (CNNs) has improved the performance of stereo-matching algorithms~\cite{gcnet,khamis2018stereonet,dispnetc,CPSN_stereo,zhang2021ednet} significantly. Typical CNN-based methods commonly rely on a cost volume, which is constructed with a predetermined search range to evaluate the matching similarity. Existing cost volume-based stereo matching can be categorized as the 3D correlation-volume-based methods~\cite{diggingNormal,wang2020fadnet,zhang2021ednet} and the 4D concatenation-volume-based methods~\cite{psmnet,gwcnet,gcnet,xu2022acvnet,ganet}. However, these methods perform poorly when applied in ill-conditioned regions like occluded regions, and texture-less regions.

The challenges associated with stereo matching in ill-conditioned regions can be simply summarized as follows: (1) Texture-less or repetitive regions show homogeneity in the RGB domain, which is difficult for CNN-based methods to extract distinguishable local matching features. 
(2) Occluded regions, which naturally lack matching correspondences and cannot be estimated by matching directly. Most methods~\cite{CSPN++,Non_Local_Proapagation,CSPN} use CNN-based spatial propagation to refine the disparity in the occluded regions using the contextual features as a guide. However, these CNN-based networks reliant on local windows exhibit a tendency to utilize the limited receptive field information from the surrounding area for disparity refinement, which leads to limited improvement in large and irregular occluded regions. Other methods in optical flow tasks like GMA~\cite{GMA_OpticalFlow} use global attention instead of local correlations for the ill-conditioned region's refinement, while uncontrolled global attention is inefficient and can even affect well-conditioned areas.

In order to improve the disparity performance in the ill-conditioned regions, in this paper, we propose to leverage restricted global spatial correlation as a guide to alleviate matching ambiguities in texture-less regions and refine the disparity in occluded regions. Our idea is that disparity within a bounded region~(e.g. an object) should be continuous. To realize this, we propose the \underline{G}lobal \underline{O}cclusion-\underline{A}ware \underline{T}ransformer (\textit{GOAT}) which introduces Vision Transformer \cite{vit_orgin} and attention mechanism to establish restricted global spatial correlation for both the matching and disparity refinement phases. 
In \textit{GOAT}, a parallel disparity and occlusion estimation module~(\textit{PDO}) is proposed to estimate the initial disparity and the occlusion mask respectively with an adaptive global search range utilizing stacked self-cross attention layers for feature aggregation and parallel cross-attention for occlusion and disparity estimation. The most related prior work is the STTR~\cite{STTR}, however, STTR employs a shared cross-attention matrix for estimating both disparity and occlusion, which leads to a trade-off between disparity and occlusion prediction accuracy. In contrast, the proposed \textit{PDO} infers occlusion and disparity independently, eliminating any possible trade-offs between the two estimates. To further enhance the disparities in the occluded regions, an iterative occlusion-aware global aggregation module~(\textit{OGA}) is proposed to refine the disparity with a restricted focus scope of the occluded regions using global spatial correlations and context guidance.

Our main contributions lie in four folds:
\begin{itemize}
    \item We explore employing restricted global spatial correlation information for stereo-matching and propose a novel stereo-matching network named \textit{GOAT}, which enables robust disparity estimation, particularly in ill-conditioned regions.
    \item We propose a parallel disparity and occlusion estimation module (\textit{PDO}) that utilizes a parallel attention mechanism to generate both disparity and occlusion masks robustly, without mutual interference.
    \item We also propose an occlusion-aware global aggregation module (\textit{OGA)} that aggregates feature with a focus scope in occluded regions using self-attention, boosting disparity estimation in occluded areas.
    \item We conducted extensive experiments on several public datasets including  SceneFlow~\cite{dispnetc}, FallingThings~\cite{Fallingthings3D}, KITTI 2015~\cite{Kitti2015}, and Middlebury~\cite{middlebury2014}. Experimental results reveal that the proposed method achieves outstanding performance on several benchmark datasets, especially in the ill-conditioned occluded regions. 
    
\end{itemize}

\begin{figure*}
	\centering	\includegraphics[width=0.88\linewidth]{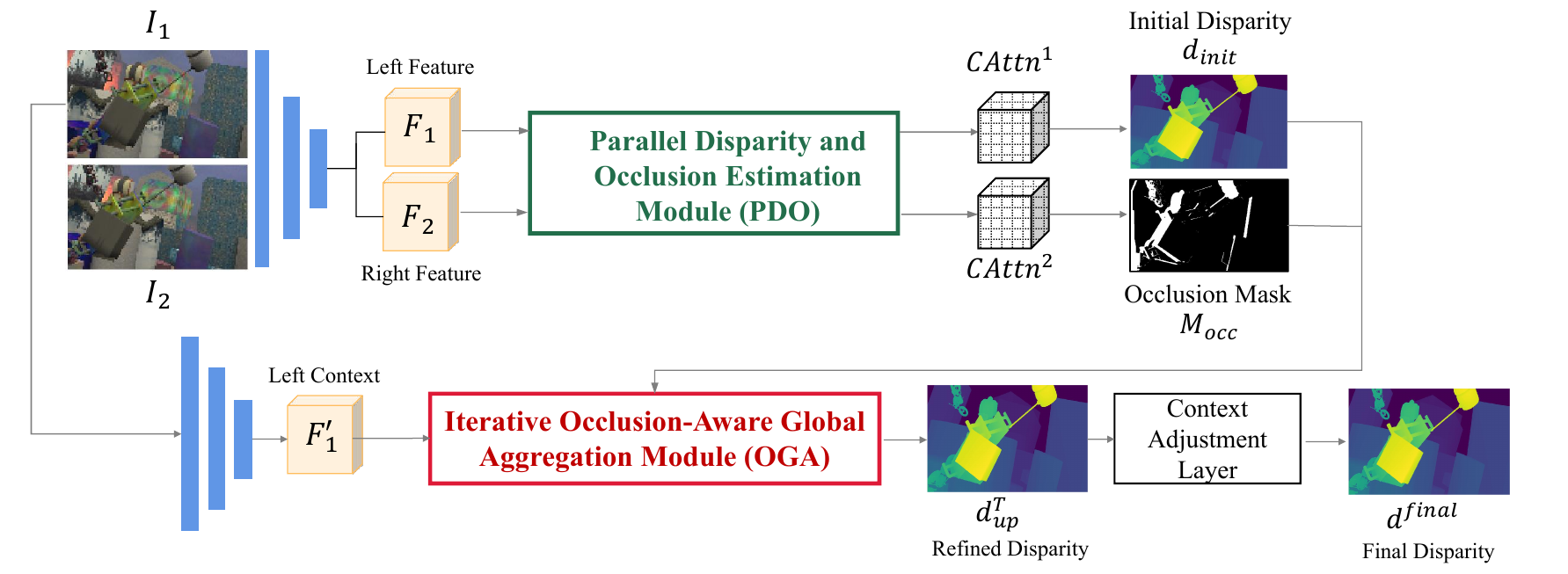}
\setlength{\abovecaptionskip}{3mm}
	\caption{Overall architecture of Global Occlusion-Aware Transformer (GOAT). }
	\label{fig:architecture}
	\vspace{-1.0 em}
\end{figure*}

\section{Related Works}
\noindent\textbf{Cost-Volume-based Methods.}\quad Pioneer work DispNetC \cite{dispnetc} utilizes a correlation layer to calculate the inner product of the left and right features at each disparity level for measuring the similarity. Although correlation volume has been proven to be effective and efficient, the loss of context information during correlation limits the ultimate performance of stereo-matching. GCNet \cite{gcnet} firstly employs the concatenation of left and right features to construct a 4D volume that encodes abundant content information for similarity measurement. The concatenation volume following stacked 3D convolution networks for aggregation is widely used in most latest state-of-the-art works including \cite{shen2021cfnet,CPSN_stereo,ganet}. In order to combine the advantages of the correlation volume and the concatenation volume, GwcNet \cite{gwcnet} adopts a group-wise correlation method to combine the correlation volume and the concatenation volume. Later work such as PCWNet~\cite{PCW-Net} follows the same architecture and exploits multi-scale volumes fusion to extract domain-invariant features, which leads to better performance.

\noindent\textbf{Guidance-Incorporated Stereo Matching.} Besides, depending on image similarity for stereo matching, some other methods utilize extra guidance information to improve stereo matching and achieve exceptional performance. Xiao et al propose a multi-task network called EdgeStereo~\cite{edgestereo} by applying a disparity-edge joint learning framework to leverage edge maps as the guidance for disparity refinement. Wu \textit{et al.}~\cite{semantic_stereo} employ semantic guidance by introducing a designed pyramid of cost volumes for describing semantic and spatial information on multiple levels. Liu \textit{et al.}~\cite{diggingNormal} propose a normal incorporated joint learning framework to explicitly leverage the surface normal as an intuitive geometric guidance to refine the ill-conditioned regions with the surface normal affinities. Although stereo-matching with guidance information is able to introduce prior knowledge beyond RGB clues for robust stereo-matching, the implementation of these approaches requires a joint-learning framework with additional supervision, which may increase the complexity and training cost of the network.\\
\noindent\textbf{Attention Mechanism in Stereo Matching.}
Recently, attention mechanisms have been introduced in the stereo-matching task to improve the quality of disparity estimation. Many works~\cite{Attention_aware_feature_aggregation,Attention_aggregation_encoder_decoder,Attention_guided} use 2D attention block for left-right feature aggregation to adaptively calibrate weight response, improving the robustness of the feature representation. Zhang \textit{et al.}~\cite{zhang2021ednet} use a warped photometric error to generate a spatial attention mask for disparity residual estimation which accelerates the training process. ACVNet~\cite{xu2022acvnet} learns an attention map from the correlation volume to suppress redundant information and enhance matching-related information in the concatenation volume. Besides, other works use an attention mechanism to replace the conventional cost volume for left-right image matching. STTR~\cite{STTR} takes the first attempt to use alternating self-cross attention modules to estimate the disparity and corresponding occlusion mask from an aspect of the transformer. GMStereo~\cite{GMStereo} presents a unified formulation using a cross-attention mechanism for three motion and 3D perception tasks: optical flow, rectified stereo matching, and unrectified stereo depth estimation from posed images. 
\section{Proposed Method}
In this section, we provide a comprehensive introduction to our proposed \underline{G}lobal \underline{O}cclusion-\underline{A}ware Stereo \underline{T}ransformer (\textit{GOAT}). The overall architecture of the proposed work is presented in Subsection \ref{Network_architecture_Oagnet}, with detailed descriptions of the proposed two specific modules provided in Subsections \ref{sec_PSAC} and \ref{sec_OAG}. The training mechanism and loss function are expounded upon in Subsection \ref{sec_supervision}.
\subsection{Overall Network Architecture}\label{Network_architecture_Oagnet}
\begin{figure*}[!t]
        \centering
\includegraphics[width=0.87\linewidth]{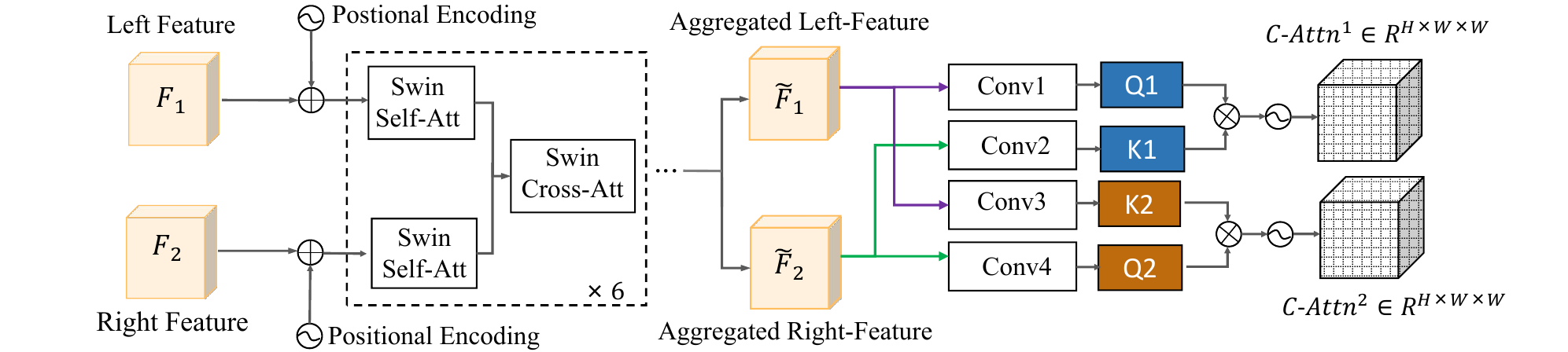}
	\caption{Parallel Disparity and Occlusion Estimation
Module Architecture. (PDO)}
	\label{fig:PSCA_fig}
	\vspace{-3mm}
\end{figure*}
The overall architecture of the proposed \textit{GOAT} is shown in Figure \ref{fig:architecture}. We decouple the stereo-matching process into matching for non-occluded regions and disparity refinement for occluded regions. In the matching phase, we propose a parallel disparity and occlusion estimation module (\textit{PDO}) which leverages both positional and global correlations between the left and right views to estimate initial disparity and the occlusion mask, respectively. In the refinement phase, we propose an iterative occlusion-aware global aggregation module (\textit{OGA}) using restricted global correlation with occlusion guidance to optimize the disparity within the occluded regions. Finally, a context adjustment layer is employed to refine the disparity from a mono-depth aspect.
\subsection{Parallel Disparity and Occlusion Estimation Module (PDO)}\label{sec_PSAC} 
Instead of using a cost volume with a predetermined search, we proposed a global-attention-based module named \textit{PDO} to compute the initial disparity and the occlusion mask. 
As illustrated in Figure \ref{fig:PSCA_fig}, After obtaining the  $F_1$ and $F_2$ $\in\mathbb{R}^{H\times W \times C}$ from the shared image extractor, we follow the architecture in~\cite{LoFTR} by introducing a self-cross alternating module to extract global context information and position bias, where the Swin-Transformers Blocks~\cite{SwinTransfomrer} with a window size of [h/2,w/2] are utilized for efficient feature aggregation. The self-cross attention module can be described as follows:
\begin{small}
\vspace{-2mm}
\begin{gather}
\setlength{\abovedisplayskip}{1pt}
F_l =softmax(\frac{Q_lK_l^{T}}{\sqrt{C}})V_l, F_r =softmax(\frac{Q_rK_r^{T}}{\sqrt{C}})V_r,  \notag  \\  
F_l =softmax(\frac{Q_rK_l^{T}}{\sqrt{C}})V_l, F_r =softmax(\frac{Q_lK_r^{T}}{\sqrt{C}})V_r,
\end{gather}
\end{small}where the first row represents the self-attentions of the left feature and right feature, while the second row represents the cross-attentions between two views. $Q$, $K$, and $V$ are obtained using a shared-weight linear projection layer with absolute positional encoding to indicate the position information.
The alternating self-cross attention modules use the global receptive field to fully aggregate the information of the left and right views, resulting in more representative and distinguishable features. In addition, positional encoding helps to constrain the aggregation range and prevent aggregating features from distant and unrelated regions with similar textures.
Once we obtained the aggregated left and right features, a parallel cross-attention module was applied to estimate the initial disparity and the occlusion mask. As illustrated in Figure \ref{fig:PSCA_fig}, we conduct parallel cross-attention between the left feature and right feature and get two cross-attention matrices $CAttn^{1}$ $\in\mathbb{R}^{H\times W \times W}$ and $CAttn^{2}$ $\in\mathbb{R}^{H\times W \times W}$. Since the normalized cross-attention reflects the similarity of left and right features, the  $CAttn^{1}$ can be regarded as a cost volume with a global search range. Besides, since occluded regions lack a corresponding pixel in the other image view, the summation of attention values of potential matching pixels for occlusion regions in $CAttn^{2}$ should yield a low response. 
According to the characteristics of these two cross-attentions, we compute the initial disparity and the occlusion mask:
\begin{gather}
\text{disp}{(i,j)} = \text{Coord}{x}^{L_{(i,j)}} - CAttn^{1} \otimes \text{RC}, \notag \\ 
\text{occlusion}_{(i,j)} = \text{sigmoid}(f_{\theta}(\sum_{k=1}^{W}CAttn^{2}_{(i,k,j)})), 
\end{gather}
where $\text{Coord}{x}^{L}{(i,j)} \in \mathbb{R}^{H \times W \times 1}$ is the standard coordinate of the left image in the horizontal direction, $\text{RC} = [0, 1, \ldots, W-1]^{T}$ is the range of all potential corresponding coordinates in the right image, and $\otimes$ denotes matrix multiplication. $f_{\theta}$ represents a small network that takes the summation of attention values of all potential matching points in $CAttn^{2}$ as input to regress the occlusion mask.

One related work is \cite{parallax}, which utilizes features extracted by a CNN for cross-attention to obtain the matching matrix for unsupervised stereo matching. However, it lacks the global context and positional encoding information introduced by alternating self-cross attention. As a result, the proposed \textit{PDO} module is more powerful in modeling texture-less and occluded regions compared to \cite{parallax}.
\begin{figure*}
	\centering
	\includegraphics[width=0.90\linewidth]{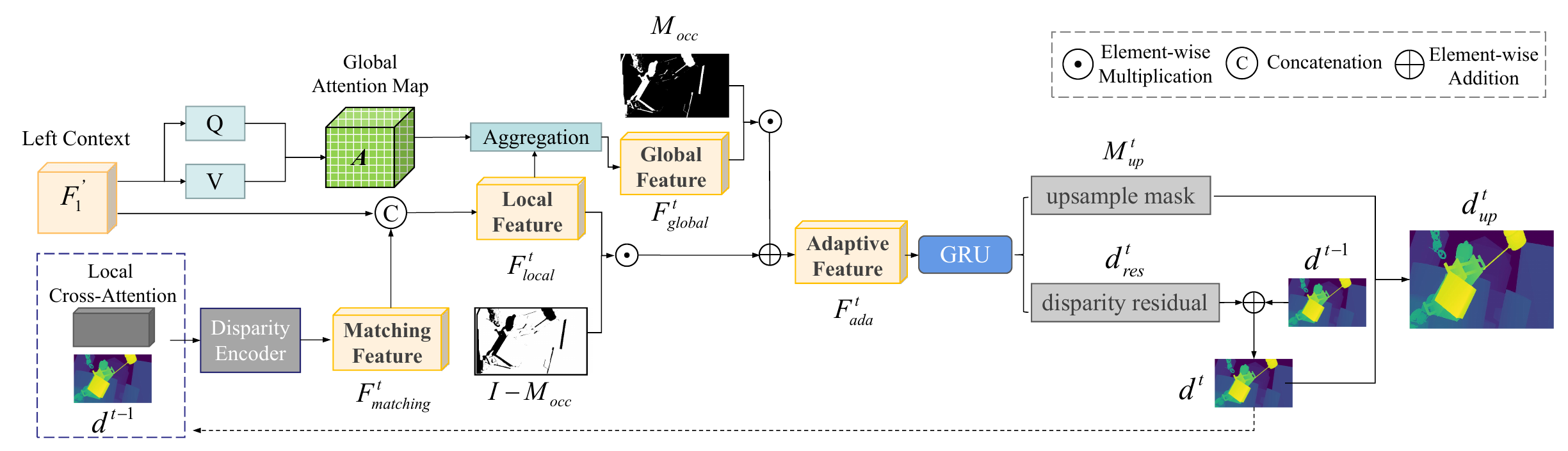}
	\caption{Iterative Occlusion-Aware Global Aggregation Module (OGA).}
	\label{fig:OAGModule}
	\vspace{-0.8 em}
\end{figure*}
\subsection{Iterative Occlusion-Aware Global Aggregation Module (OGA)}\label{sec_OAG}
After obtaining the initial disparity and occlusion mask at low resolution, the disparities in ill-conditioned regions, such as occluded areas, remain problematic, since they are difficult to estimate accurately via matching alone. To further enhance the disparity estimation performance, we propose an iterative refinement module based on self-attention, namely \textit{OGA} module, which aggregates features from valid non-occluded regions into invalid occluded regions using global spatial correlation. Similar to RAFT~\cite{RAFT}, a convex upsampling layer is used to upsample the disparity to a higher resolution. The overall structure of the \textit{OGA} module is shown in Figure \ref{fig:OAGModule}.

The input of the \textit{OGA} module is the disparity $d^{t-1}$ of stage $t-1$ as well as the left context $F_1'$ extracted from a CNN. We also construct a local cross-attention that measures the similarity between the left and right features around the $d^{t-1}$ with a search range of $r$ by sampling from the cross-attention matrix $CAttn^{1}$ in the \textit{PDO} module. The current disparity $d^{t-1}$ and its corresponding local cross-attention are then passed to a disparity encoder to obtain the matching feature $F_{matching}^t$. Meanwhile, the left context $F_1'$ is further concatenated with $F_{matching}^t$ to supplement local feature $F_{local}^t$ from a mono-depth aspect. Such information is sufficient for disparity optimization in the non-occluded regions. As for occluded regions, we calculate the global spatial correlation of the left image through the self-attention module and obtain a self-attention matrix $A \in \mathbb{R}^{H\times W \times H \times W}$. For arbitrary specific point $(i,j)$, we obtain its correlation with all other pixels in the left view by consulting the attention map $A_{i,j} \in \mathbb{R}^{H \times W}$. Then, we perform feature aggregation to derive global feature $F_{global}^t$. With local feature $F_{local}^t$ and global feature $F_{global}^t$ obtained, we then adopt an occlusion-aware global aggregation mechanism as shown in Figure~\ref{fig:OAGModule}. We reserve the local feature at the non-occluded region and keep the global feature at the occluded region to generate an adaptive feature $F_{ada}^t$ for overall disparity refinement. On the one hand, local features are sufficient for non-occluded regions to perform disparity refinement. On the other hand, we can prevent the local features of occluded regions, which are less confident because of the matching ambiguity, from propagating to non-occluded regions through the attention map like ~\cite{GMA_OpticalFlow}. This can effectively reduce the degradation of features. Therefore, the proposed \textit{OGA} module can make good use of the global spatial correlations at the ill-conditioned regions as well as avoid harmful propagation.
The whole process can be described as follows:
\begin{gather}
   F_{ada}^{t} = A \otimes F_{local}^{t} \odot M_{occ} + F_{global}^{t} \odot (I-M_{occ}), \\ \notag
   F_{local}^{t} = concat(F_{matching}^t,F_{1}^{'}),
\end{gather}
where $M_{occ}$ indicates the occlusion mask, $I$ is an identity matrix, and $\odot$ denotes element-wise multiplication.
After feature aggregation, we employ a GRU~\cite{GRU} unit to regress the disparity residual and an upsample mask, where we compute the disparity $d^t$ at the current iteration and use the upsample mask to increase the resolution:
\begin{gather}
       d^{t}_{res},M_{up}^{t} = GRU(F_{ada}^{t}),  \notag \\
   d^{t} = max(0,d^{t}_{res}+d^{t-1}), \\
   d^{t}_{up} =  d^{t} \ast M_{up}^{t}, \notag
\end{gather}
where the $M_{up}^{t} \in\mathbb{R}^{H \times W \times S \times S}$, $S$ is the upsample scale, and $\ast$ denotes convolution. The upsampled disparity $d^{T}_{up}$ of the last iteration $T$ is further passed to a context adjustment layer~\cite{STTR} to derive final disparity $d^{final}$, which recovers fine-grained disparity details from a mono-depth aspect. This layer utilizes the left image and the current disparity map to regress the disparity residual.
\subsection{Occlusion and Disparity Supervision}\label{sec_supervision}
We supervised the network with groundtruth disparity and occlusion mask. Since the \textit{GOAT} is an iterative network, we follow the sequence loss proposed in~\cite{RAFT} to supervise the disparity at different iterations, which is the $l_1$ distance between the ground truth disparity and the estimated disparity at each iteration with exponentially increasing weights. The loss can be defined as follows: 
\begin{small}
\begin{gather}
L_{disp} = \sum_{i=0}^{T} \gamma^{T-t} \left\|d^{gt}-d^{t}_{up}\right\| + \left\|d^{gt}-d^{final}\right\|,
\end{gather} 
\end{small}where the $T$ is the iteration number which in our case equals 12 and set the increasing weight $\gamma$ to 0.95. 
For occlusion supervision, the cross-entropy loss is deployed for effective training:
\begin{small}
\begin{gather}
L_{occ} = -\frac{1}{2}\sum_{i}^{2}(O_{gt}\log(O_{i}) +(1-O_{gt})\log(1-O_{i})).
\end{gather} 
\end{small}
The final loss is the weight summation of disparity loss and occlusion loss.
\begin{gather}
L_{total} = \lambda_{1} \times L_{disp} + \lambda_{2} \times L_{occ} .
\end{gather}

\section{Experimental Results}
\begin{table*}[!ht]
\setlength{\tabcolsep}{1.0mm}
\centering
\caption{Ablation study of our proposed \textit{GOAT} network on the SceneFlow dataset. We conduct ablation studies on the proposed \textit{PDO} and \textit{OGA} modules. As well as compared with other attention-based disparity estimation and refinement modules like \textit{STTR} \cite{STTR} and \textit{GMA} \cite{GMA_OpticalFlow}. The '*' represents a higher resolution. We calculated the EPE and P1(outliers) both in the overall and the occluded regions separately.}
\scalebox{0.90}{

\begin{tabular}{ll|ccc|ccc|c|c|cc|cc|c}
\hline
\multicolumn{2}{c|}{\multirow{3}{*}{Method}} &
  \multicolumn{3}{c|}{\multirow{2}{*}{\begin{tabular}[c]{@{}c@{}}Disparity\\ Estimation\end{tabular}}} &
  \multicolumn{3}{c|}{\multirow{2}{*}{\begin{tabular}[c]{@{}c@{}}Update\\ Module\end{tabular}}} &
   &
   &
  \multicolumn{2}{c|}{\multirow{2}{*}{EPE}} &
  \multicolumn{2}{c|}{\multirow{2}{*}{P1(\%)}} &
  \multirow{3}{*}{\begin{tabular}[c]{@{}c@{}}Occ\\ mIOU\end{tabular}} \\
\multicolumn{2}{c|}{} &
  \multicolumn{3}{c|}{} &
  \multicolumn{3}{c|}{} &
  \multirow{2}{*}{\begin{tabular}[c]{@{}c@{}}\textbf{CA}\\ \textbf{Layer}\end{tabular}} &
   &
  \multicolumn{2}{c|}{} &
  \multicolumn{2}{c|}{} &
   \\ \cline{3-8} \cline{11-14}
\multicolumn{2}{c|}{} &
  \begin{tabular}[c]{@{}c@{}}Cost\\ Volume\end{tabular} &
  STTR &
  \textbf{PDO} &
  RAFT &
  GMA &
  \textbf{OGA} &
   &
   &
  All &
  Occ &
  All &
  Occ &
   \\ \cline{1-9} \cline{11-15} 
\multicolumn{2}{l|}{Baseline} &
  \checkmark &
   &
   &
  \checkmark &
   &
   &
   &
   &
  0.79 &
  2.27 &
  9.2\% &
  25.6\% &
  - \\ \cline{1-9} \cline{11-15} 
\multicolumn{2}{l|}{STTR} &
   &
  \checkmark &
   &
  \checkmark &
   &
   &
   &
   &
  0.78 &
  2.31 &
  10.0\% &
  28.4\% &
  0.81 \\
\multicolumn{2}{l|}{PDO} &
   &
   &
  \checkmark &
  \checkmark &
   &
   &
   &
   &
  0.65 &
  1.96 &
  7.2\% &
  22.2\% &
  0.83 \\ \cline{1-9} \cline{11-15} 
\multicolumn{2}{l|}{PDO~+~GMA} &
   &
   &
  \checkmark &
   &
  \checkmark &
   &
   &
   &
  0.62 &
  1.86 &
  7.0\% &
  21.9\% &
  0.83 \\
\multicolumn{2}{l|}{PDO~+~OGA} &
   &
   &
  \checkmark &
   &
   &
  \checkmark &
   &
   &
  0.57 &
  1.78 &
  6.7\% &
  20.9\% &
  0.83 \\ \cline{1-9} \cline{11-15} 
\multicolumn{2}{l|}{PDO~+~OGA~+~CA(Full)} &
   &
   &
  \checkmark &
   &
   &
  \checkmark &
  \checkmark &
   &
  0.55 &
  1.72 &
  6.6\% &
  19.9\% &
  0.83 \\
\multicolumn{2}{l|}{PDO~+~OGA~+~CA*(Full)} &
   &
   &
  \checkmark &
   &
   &
  \checkmark &
  \checkmark &
   &
  0.47 &
  1.53 &
  5.6\% &
  18.6\% &
  0.94 \\ \hline
\end{tabular}

}

\label{tab:OAGNet_ablation}
 \vspace{-0.8em} 
\end{table*}
\begin{figure*}[!htbp]
\setlength{\abovecaptionskip}{0.cm}
	\centering
\includegraphics[width=1.0\linewidth]{ 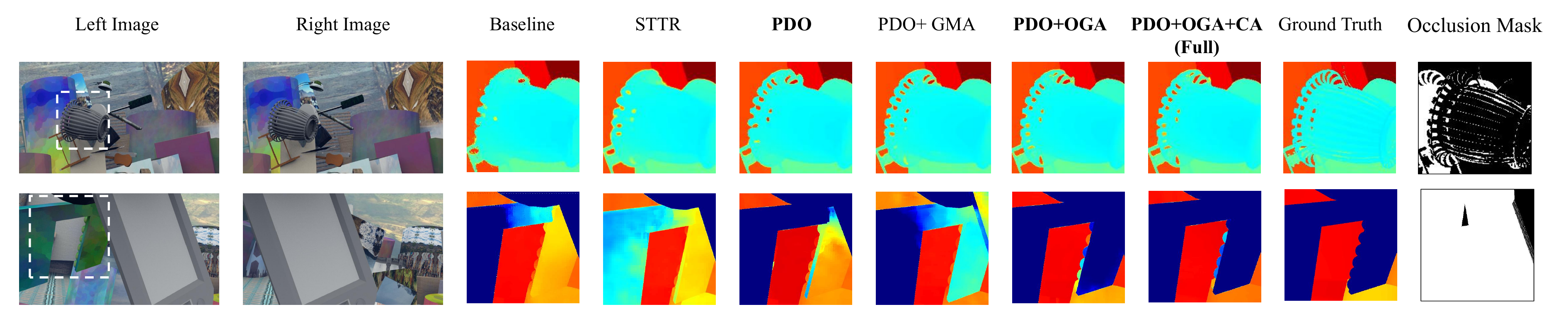}
	\caption{Visualizations of ablation study on SceneFlow dataset. We cropped and enlarged the selected part of the disparity map for easier viewing.}
	\label{fig:OAG_Ablation_VIS}
	\vspace{-0.8 em}
\end{figure*}
\subsection{Datasets}
We evaluate our method on multiple public benchmark datasets including SceneFlow~\cite{dispnetc}, Falling  Things~\cite{Fallingthings3D} KITTI 2015~\cite{Kitti2015}, and Middlebury~\cite{middlebury2014}. As the proposed network requires ground-truth occlusion masks for training, which are not provided in the several datasets, we generate the ground-truth occlusion masks using left-right consistency. More details can be seen in our \textit{supplementary material}. The SceneFlow dataset is a synthetic dataset containing 39,823 stereo image pairs with random flying objects. The Falling Things dataset is another synthetic dataset with more realistic indoor scenes. The KITTI 2015 dataset comprises real-world scenes that have sparse ground-truth disparity captured using LiDAR. For the Middlebury dataset, the evaluation is conducted using the standard Middlebury Stereo Evaluation-Version 3.
\subsection{Implementation Details}
We implemented our \textit{GOAT} network by PyTorch trained with 4 NVIDIA 3090 GPUs. For the SceneFlow dataset, we trained the networks for 80 epochs using a batch size of 8 with an initial learning rate of 4e-4 following a step learning rate decay strategy. For the Falling Things dataset, we trained for 10 epochs with a constant learning rate of 4e-4.
Compared to SceneFlow dataset, the Falling Things dataset~\cite{Fallingthings3D} has enhanced scene realism and better semantics in occluded region, therefore we use it for more comprehensive ablation studies. For both above dataset, we randomly cropped the input images to 320$\times$640. 
For the KITTI 2015 dataset, we fine-tune our networks with the SceneFlow pre-trained model. Mixed datasets of KITTI 2012 and KITTI 2015, totaling 400 image pairs, were used for the initial 400 epochs with a random crop size of 320×1088. The model with the best validation performance was chosen, followed by another $200$ epochs of fine-tuning on the KITTI 2015 training set to obtain the final model.
For the Middlebury dataset with only 23 images, we first evaluated generalization on the Middlebury training set using the SceneFlow pre-trained model, then fine-tuned it at half-resolution for benchmark assessment. Please refer to the \textit{supplementary material} for more training details.
\begin{table*}[!t]
\setlength{\tabcolsep}{1.2mm}
    \centering
\caption{Quantitative comparison of \textit{GOAT} and other methods on the SceneFlow. We adopt the EPE-All results from the original papers. Due to incomplete disparity evaluation of the occluded regions in some works, we calculate EPE-Occ using the corresponding official pre-trained models. Proposed \textit{GOAT} ranks top for overall and occluded regions.$\textbf{\textcolor[rgb]{1,0,0}{~Red Bold:Best}}$. \textbf{Bold:Second}.}
\resizebox{\textwidth}{!}{
    \begin{tabular}{c|c|c|c|c|c|c|c|c} \hline
         Model & PSMNet~\cite{psmnet} & AANet++~\cite{AANet} & RaftStereo~\cite{RaftStereo} & PCW-Net~\cite{PCW-Net} & STTR-light~\cite{STTR} & ACVNet~\cite{xu2022acvnet} & 
         IGEVStereo~\cite{IGEV-Stereo}&
         \textbf{GOAT (Ours)}\\ \hline
         EPE-All & 1.09 &0.72 & 0.69 & 0.86 & 4.14 & \textbf{0.48} &\textbf{\textcolor[rgb]{1,0,0}{0.47}}& \textbf{\textcolor[rgb]{1,0,0}{0.47}} \\ 
         EPE-Occ & 3.14 &  2.44 & 2.14 & 2.54 & 23.9 & 1.65 & \textbf{1.61}& \textbf{\textcolor[rgb]{1,0,0}{1.53}} \\ \hline
    \end{tabular}}
    \label{tab:sceneflow_performance}
\vspace{-5mm}
\end{table*}
\subsection{Ablation Studies}
We conducted ablation studies on the SceneFlow and Falling Things datasets. We report the standard end-point error (EPE) and P1-value (outliers) for overall regions~(All) and occluded regions~(Occ), respectively. For occlusion mask evaluation, we compute the mean Intersection over Union (mIoU) between the ground truth and the predicted occlusion mask. The relevant results  of the Sceneflow dataset are shown in Table \ref{tab:OAGNet_ablation}, where we use a simplified version of \cite{RaftStereo} as the Baseline. For more ablation studies in the FallingThings Datset, please refer to our \textit{supplemenatry aterials}\\
\textbf{Parallel Disparity and Occlusion Estimation Module (PDO):} As depicted by Table \ref{tab:OAGNet_ablation}, compared with the Baseline integrating the \textit{PDO} module (designated as PDO) exhibits a remarkable improvement in terms of EPE for both overall and occluded regions. We also compared our proposed \textit{PDO} modules with another transformer-based method  by replacing the PDO module with a disparity estimation module proposed in STTR~\cite{STTR}. As demonstrated in Table \ref{tab:OAGNet_ablation}, our \textit{PDO} shows better disparity estimation performance with smaller errors, especially in the occluded regions, where the STTR-based method reveals even bigger EPE errors than the baseline. Further insight into the efficacy of the \textit{PDO} module can be gained from the 1st row of Figure \ref{fig:OAG_Ablation_VIS}, which demonstrates that the PDO derives a more accurate structural representation of the object compared with Baseline and \textit{STTR}, as \textit{PDO} module reduces the matching ambiguity when dealing with the texture-less and occluded regions. \\
\textbf{Iterative Occlusion-Aware Global Aggregation Module (\textit{OGA}):} Table \ref{tab:OAGNet_ablation} illustrates the effectiveness of the \textit{OGA} module. Model with the \textit{OGA} module, which is named as \textit{PDO+OGA} can reduce the EPE in the occluded regions from 1.96 to 1.78 in the SceneFlow dataset with an improvement of 10.1\%, which is more effective compared with naive global-attention-based \textit{GMA}\cite{GMA_OpticalFlow} module with an improvement of 5.1\%. Moreover, The \textit{OGA} module is also able to maintain the disparity at the non-occluded regions due to the restricted global attention mechanism. As depicted in Figure \ref{fig:OAG_Ablation_VIS}, the \textit{PDO+OGA} shows less error and enhanced robustness in the occluded regions (marked by white boxes) compared to the \textit{PDO} only and \textit{PDO+GMA}. Besides, it also shows better disparity estimation at the non-occluded regions while the \textit{PDO+GMA} fails to estimate well. 
Moreover, incorporating the context adjustment module into the whole, designated as \textit{PDO+OGA+CA}, results in further improved performance. \\
\noindent\textbf{Resolution:} Like \cite{RaftStereo}, we employed the \textit{PDO} and \textit{OGA} modules at both 1/8 and 1/4 resolutions. As shown in Table~\ref{tab:OAGNet_ablation}, increasing the resolution yields better performance, while consuming much bigger GPU memory for the self-attention computation. 

\begin{figure}[!t]
	\centering
\includegraphics[width=1.0\linewidth]{ 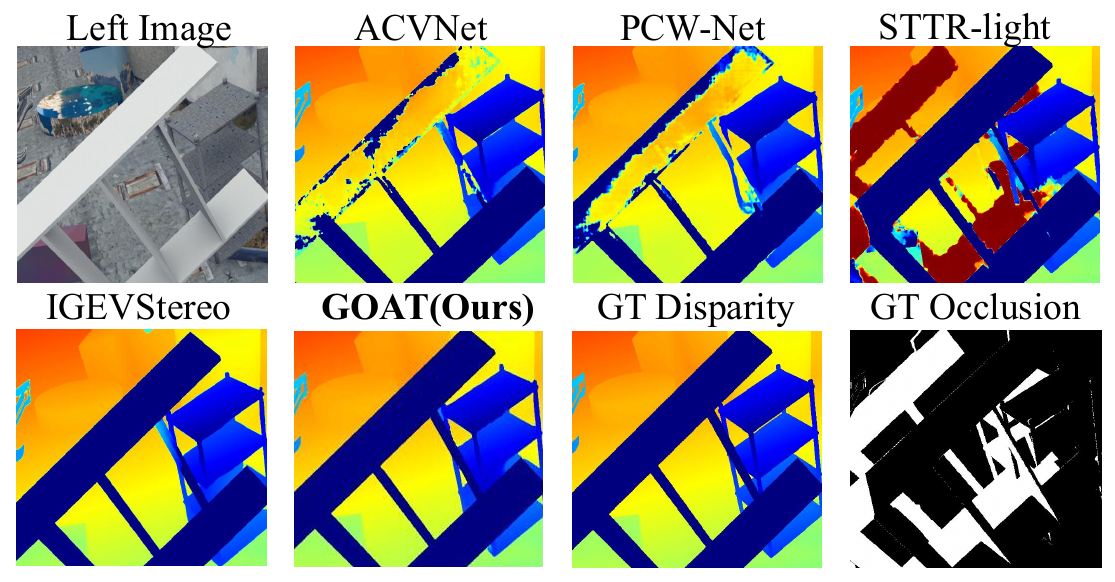}
	\caption{Qualitative comparison on SceneFlow dataset with other superior works.}
	\label{fig:compareds_sf}
	\vspace{-3mm}
\end{figure}
\subsection{Performance Evaluation}
In this subsection, we compare our method with other top-performing methods using multiple datasets. \\
\textbf{SceneFlow.} For quantitative evaluation demonstrated in Table \ref{tab:sceneflow_performance}, our proposed method ranks at the top for occluded regions, surpassing all competing methods and even very recent state-of-the-art methods such as IGEVStereo~\cite{IGEV-Stereo} and PCW-Net~\cite{PCW-Net}. Note that while IGEVStereo~\cite{IGEV-Stereo} requires 32 iterations for disparity refinement, our proposed \textit{GOAT} achieves equivalent disparity performance in overall regions with only 12 iterations, and surpasses IGEVStereo~\cite{IGEV-Stereo} in occluded regions by a large margin. This further illustrates the advantages of our proposed GOAT in optimizing disparity in the occluded regions.
\begin{table}[!t]
    \centering
        \caption{Benchmark results on KITTI 2015 test set. The \textbf{"Noc"} and \textbf{"All"} indicate the non-occluded and overall regions, respectively. The \textbf{"fg"} and \textbf{"all"} indicate the foreground and overall regions, respectively. The results report the percentage of outliers over the available ground truth disparities.}
    \scalebox{0.90}{
    \begin{tabular}{c|cc|cc|c} \hline
     \multirow{2}{*}{\quad Method  \qquad } & \multicolumn{2}{c|}{Noc (\%)} & \multicolumn{2}{c|}{All (\%)} & \multirow{2}{*}{Time (s)}\\ \cline{2-5}
     & fg & all & fg & all & \\ \hline
    GANet \cite{ganet} & 3.37 & 1.73 & 3.82 & 1.93 & 0.36 \\  
    PSMNet \cite{psmnet} & 4.31 & 2.14 & 4.62 & 2.32 & 0.41 \\ 
    GwcNet \cite{gwcnet} & 3.49 & 1.92 & 3.93 & 2.11 & 0.32 \\ 
    AANet \cite{AANet} & 4.93 & 2.32 & 5.39 & 2.55 & 0.075 \\
    DispNetC \cite{dispnetc} & 3.72 & 4.05 & 4.41 & 4.34 & 0.06 \\
    FADNet \cite{wang2020fadnet} & 3.07 & 2.59 & 3.50 & 2.82 & 0.05 \\
    IGEVStereo~\cite{IGEV-Stereo} & \textbf{2.62} & 1.49 & \textbf{2.67}& \textbf{\textcolor[rgb]{1,0,0}{1.59}} & 0.83\\
    HITNet \cite{HINET} & 2.72 & 1.74 & 3.20 & 1.98 & 0.02 \\
    LEAStereo~\cite{leastereo} & 2.65& 1.51 & 2.91 & 1.65 & 0.30 \\
    RAFTStereo~\cite{RaftStereo} & 2.94 & \textbf{1.45} & 2.94 & 1.82 & 0.38\\ 
    GMStereo~\cite{GMStereo} & 2.97 & 1.61 & 3.14 & 1.77 & 0.38\\ 
    CFNet~\cite{cfnet} & 3.25 & 1.73 & 3.56 & 1.88 & 0.38\\ 
    ACVNet~\cite{xu2022acvnet} & 2.84 & 1.52& 3.07 & \textbf{1.65} & 0.20 \\
    PCW-Net~\cite{pcw} & 2.93 & \textbf{\textcolor[rgb]{1,0,0}{1.26}} & 3.16 & 1.67 & 0.44\\ \hline
    \textbf{GOAT(Ours)}& \textbf{\textcolor[rgb]{1,0,0}{2.43}} & 1.71 & \textbf{\textcolor[rgb]{1,0,0}{2.51}} & 1.84 & 0.29 \\ \hline
    \end{tabular}}
    \label{tab:kitti_test}
    \vspace{-1.5mm}
\end{table}
\begin{figure}[!htp]
	\centering
\includegraphics[width=1.00\linewidth]{ 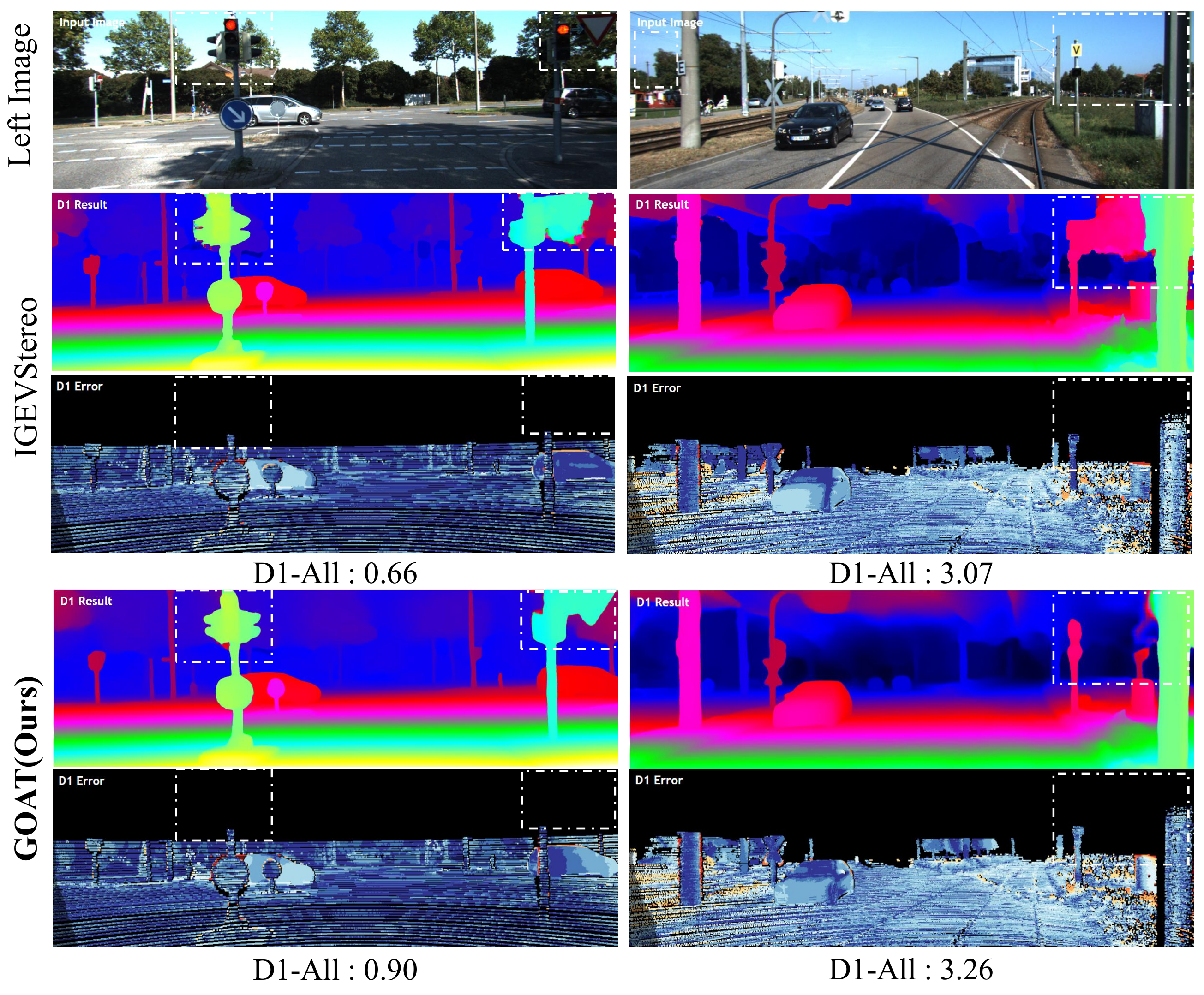}
\caption{Visualization comparison on KITTI 2015 test set between IGEVStereo~\cite{IGEV-Stereo} and our \textit{GOAT}. The $2^{nd}$ and $4^{th}$ line show estimated disparity maps, and the $3^{rd}$ and $5^{th}$ line display the corresponding errors. The error map indicates that colored regions have LiDAR annotation while black regions lack annotation, which means the \textbf{D1-All cannot fully represents the disparity estimation performance on the whole scene}. Although our model has a higher D1-All error, it exhibits improved structures and fewer artifacts in regions where the ground-truth disparity is missing.}
\label{fig:compared_with_pcwnet}
\vspace{-5 mm}
\end{figure}
\begin{figure*}[!tp]
	\centering
 \setlength{\abovecaptionskip}{1mm}
\includegraphics[width=1.0\linewidth]{ 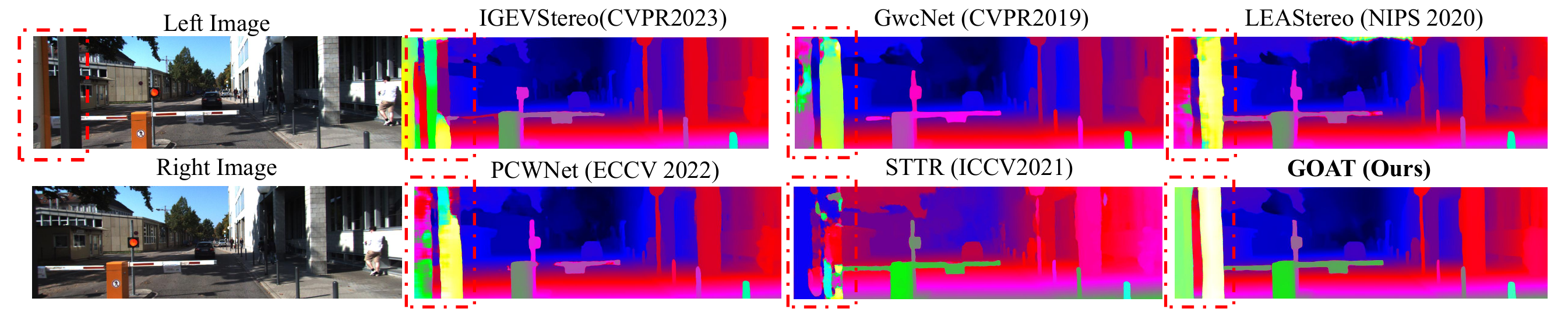}
	\caption{Performance on KITTI 2015 test set. Our method obviously exhibits better results in the severely occluded regions. }
\label{fig:compared_with_kitti_refine}
	\vspace{-0.8 em}
\end{figure*}
For qualitative evaluation shown in Figure \ref{fig:compareds_sf}, proposed \textit{GOAT} generates disparity maps with more detailed and precise structures in texture-less areas. In contrast, other methods exhibit less satisfactory performance, with missing details and artifacts. \\
\textbf{KITTI 2015.}~For KITTI dataset evaluation, we follow the standard protocol to submit our fine-tuned results to KITTI leaderboard~\cite{Kitti2015}. Table \ref{tab:kitti_test} demonstrates the evaluation performance on the KITTI 2015 test set. 
In our assessment of overall (All) regions, including occluded areas, our method distinctly excels in its performance on foreground (fg) objects with key items like cars and pedestrians, achieving a D1-Error of 2.51. The results surpass very recent methods including PCWNet~\cite{PCW-Net} and IGEVStereo~\cite{IGEV-Stereo}. Importantly, in the context of real-world autonomous driving applications, foreground regions like pedestrians and cars are of great importance, where our method proves notably proficient.
As evidenced in Figure \ref{fig:compared_with_kitti_refine}, for out-of-view regions marked by the red box which lack the corresponding pixels, our proposed \textit{GOAT} still succeeds in estimating the disparity by showing better depth consistency and clearer structures. At the same time, other methods fail to generate satisfactory results.

\begin{figure}[!ht]
	\centering
	\begin{minipage}{0.241\linewidth}
		\centering
        \setlength{\abovecaptionskip}{0.cm}
    \captionsetup{font=scriptsize}		
			\caption*{Left Image}
		\includegraphics[width=1.0\linewidth]{ 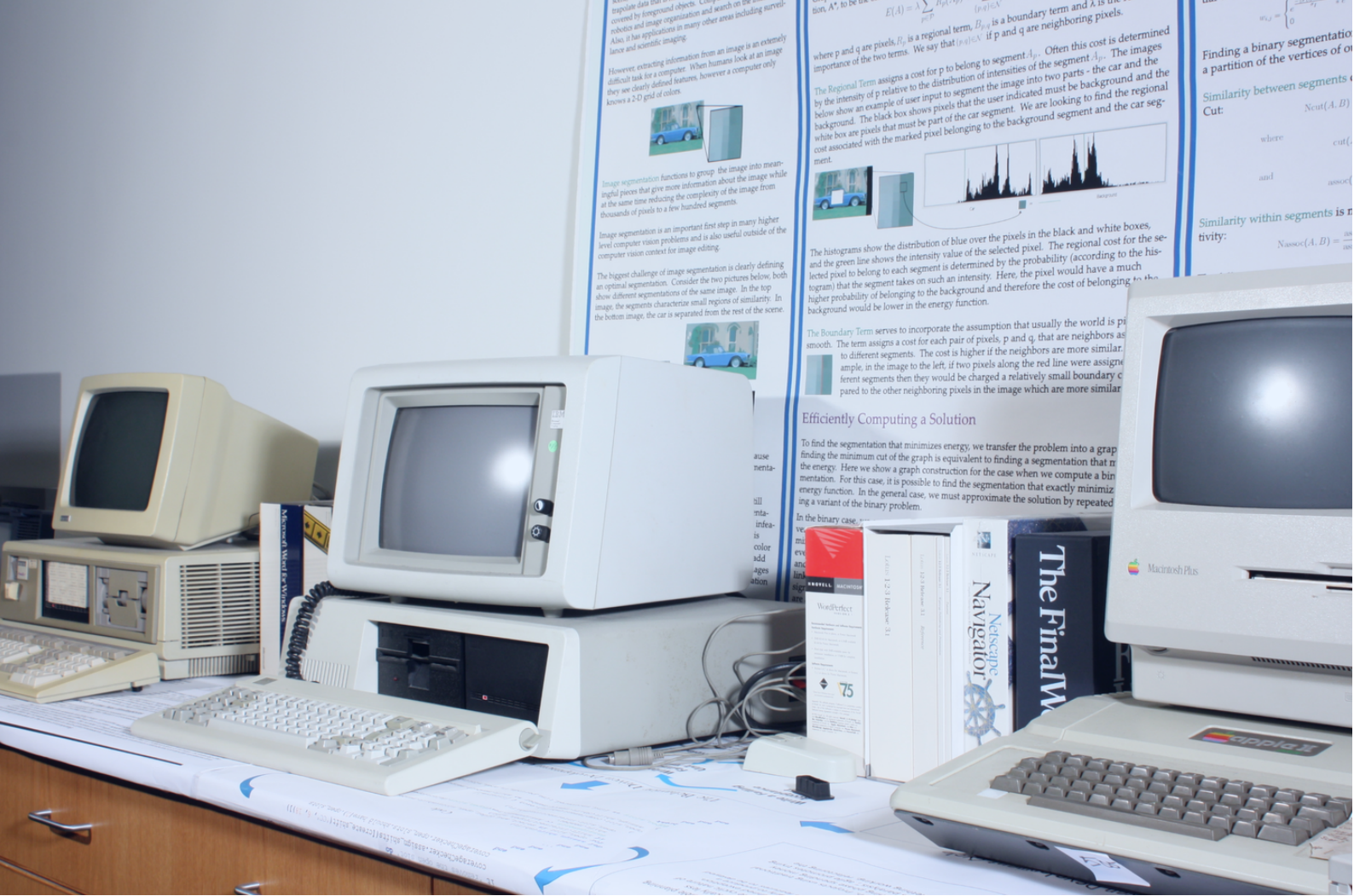}
		\label{chutian12323}
                    \vspace{-3mm}
	\end{minipage}
	\begin{minipage}{0.241\linewidth}
		\centering
        \setlength{\abovecaptionskip}{0.cm}
    \captionsetup{font=scriptsize}		
			\caption*{STTR}
		\includegraphics[width=1.0\linewidth]{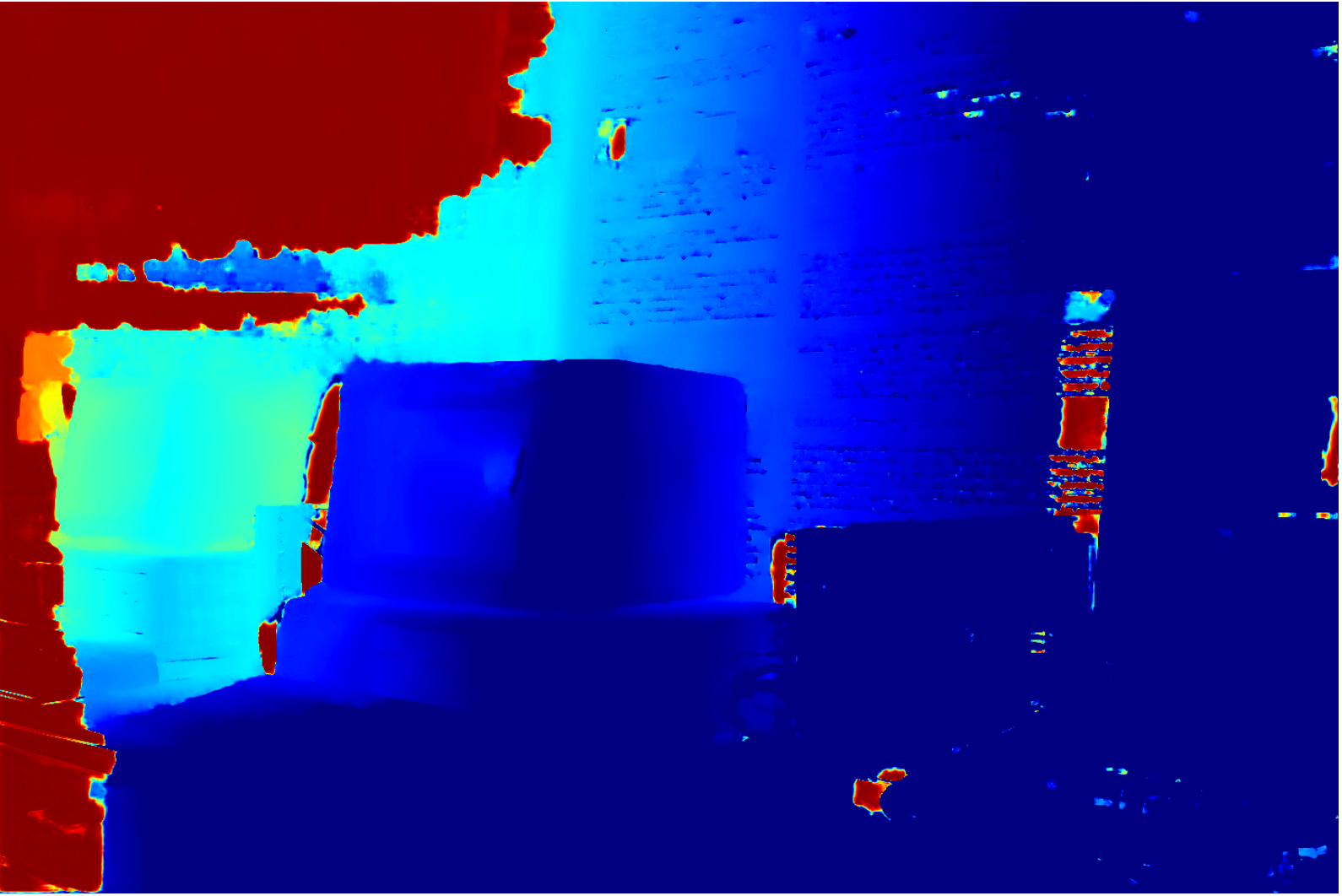}\hspace{0mm}
		\label{chutian4}
          \vspace{-3mm}
	\end{minipage}
	\begin{minipage}{0.241\linewidth}
		\centering
          \setlength{\abovecaptionskip}{0.cm}
    \captionsetup{font=scriptsize}		
			\caption*{PCWNet}
\includegraphics[width=1.0\linewidth]{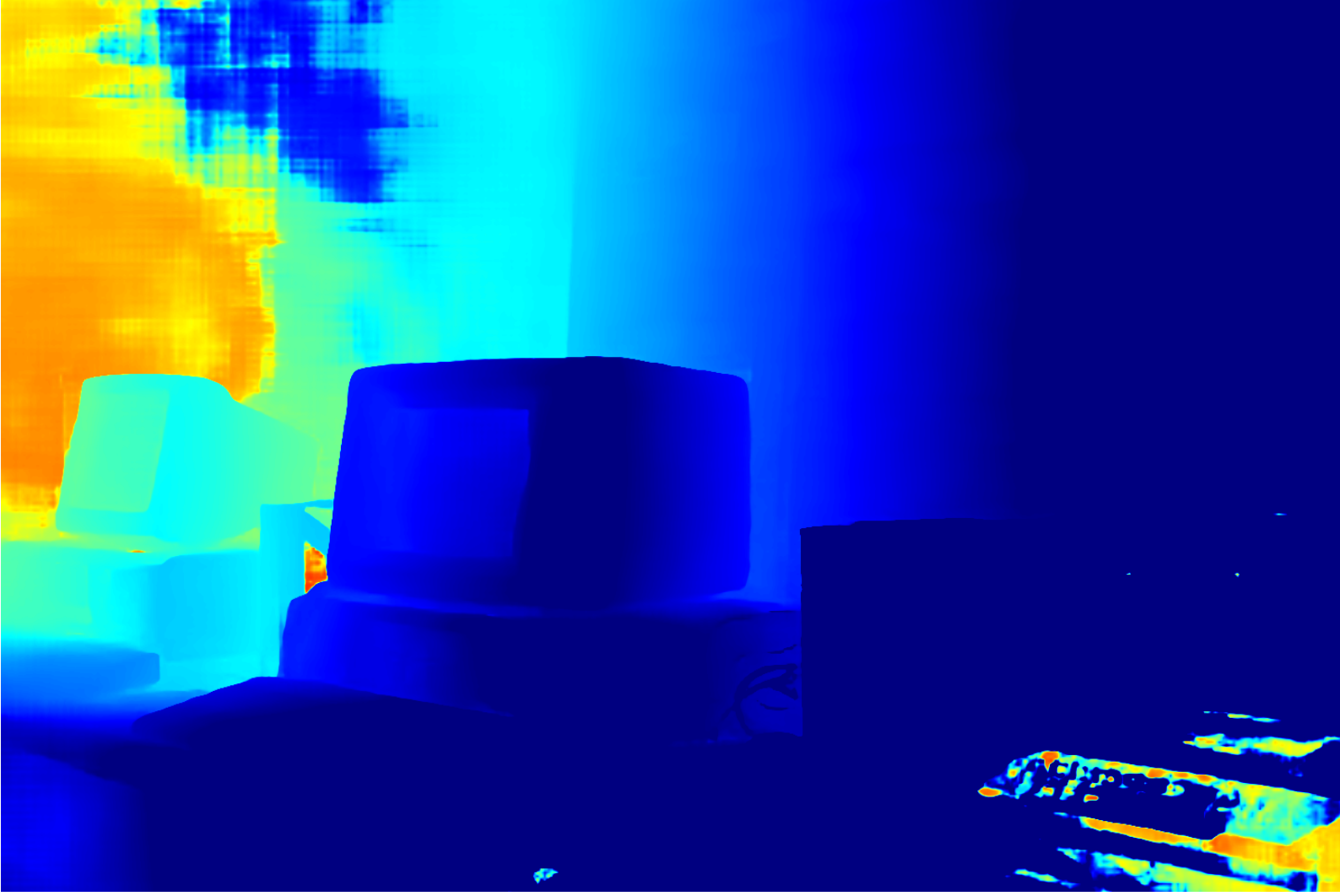}
		\label{chutian5}
                \vspace{-3mm}
	\end{minipage}
	\begin{minipage}{0.241\linewidth}
		\centering
          \setlength{\abovecaptionskip}{0.cm}
    \captionsetup{font=scriptsize}		
			\caption*{\textbf{GOAT(Ours)}}
		\includegraphics[width=1.0\linewidth]{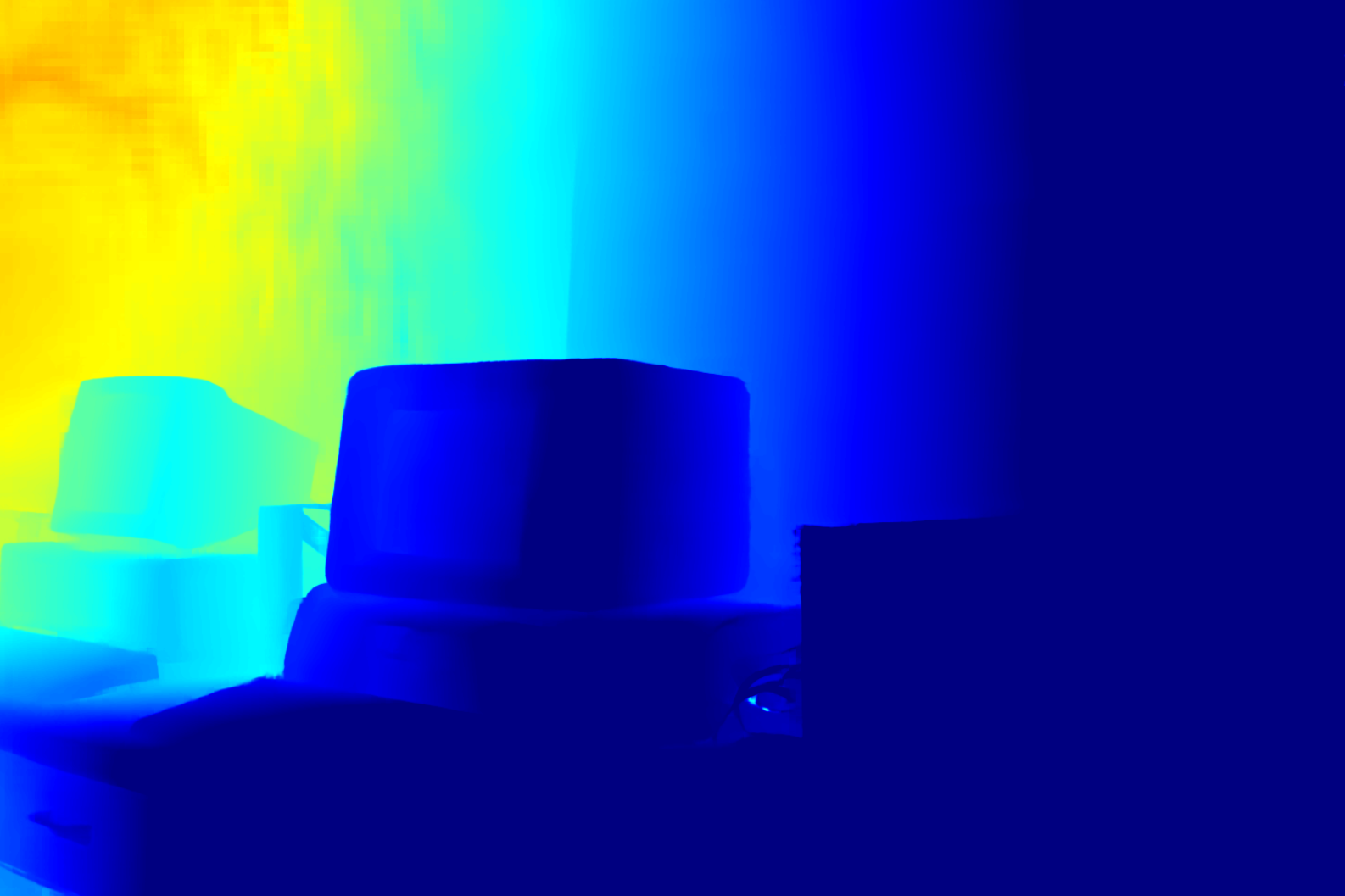}
		\label{chutian6}
                \vspace{-3mm}
	\end{minipage}
 
 	\begin{minipage}{0.241\linewidth}
		\centering
		\includegraphics[width=1.0\linewidth]{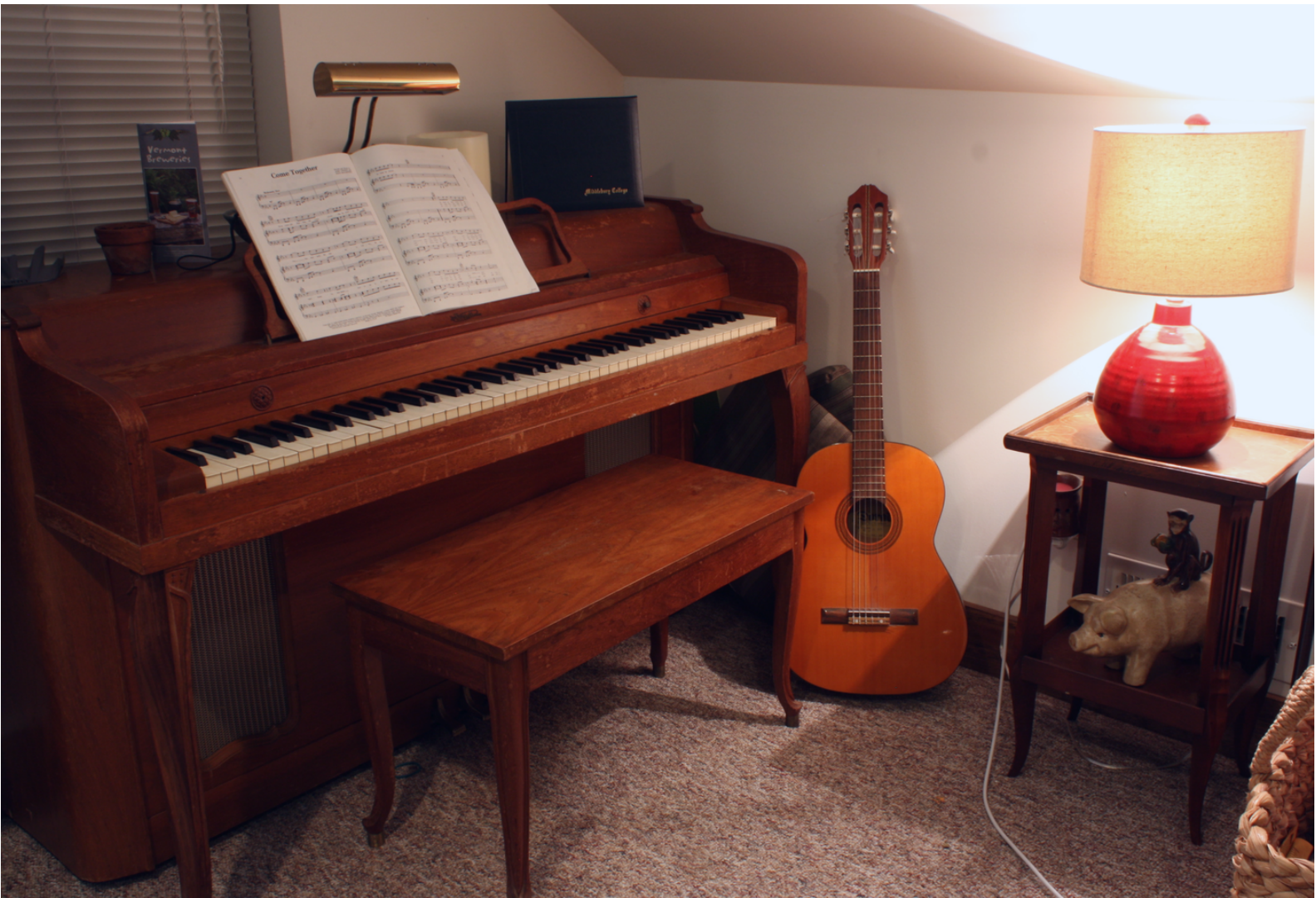}
		\label{chutian12232}
                    \vspace{-3mm}
	\end{minipage}
	\begin{minipage}{0.241\linewidth}
		\centering
		\includegraphics[width=1.0\linewidth]{ 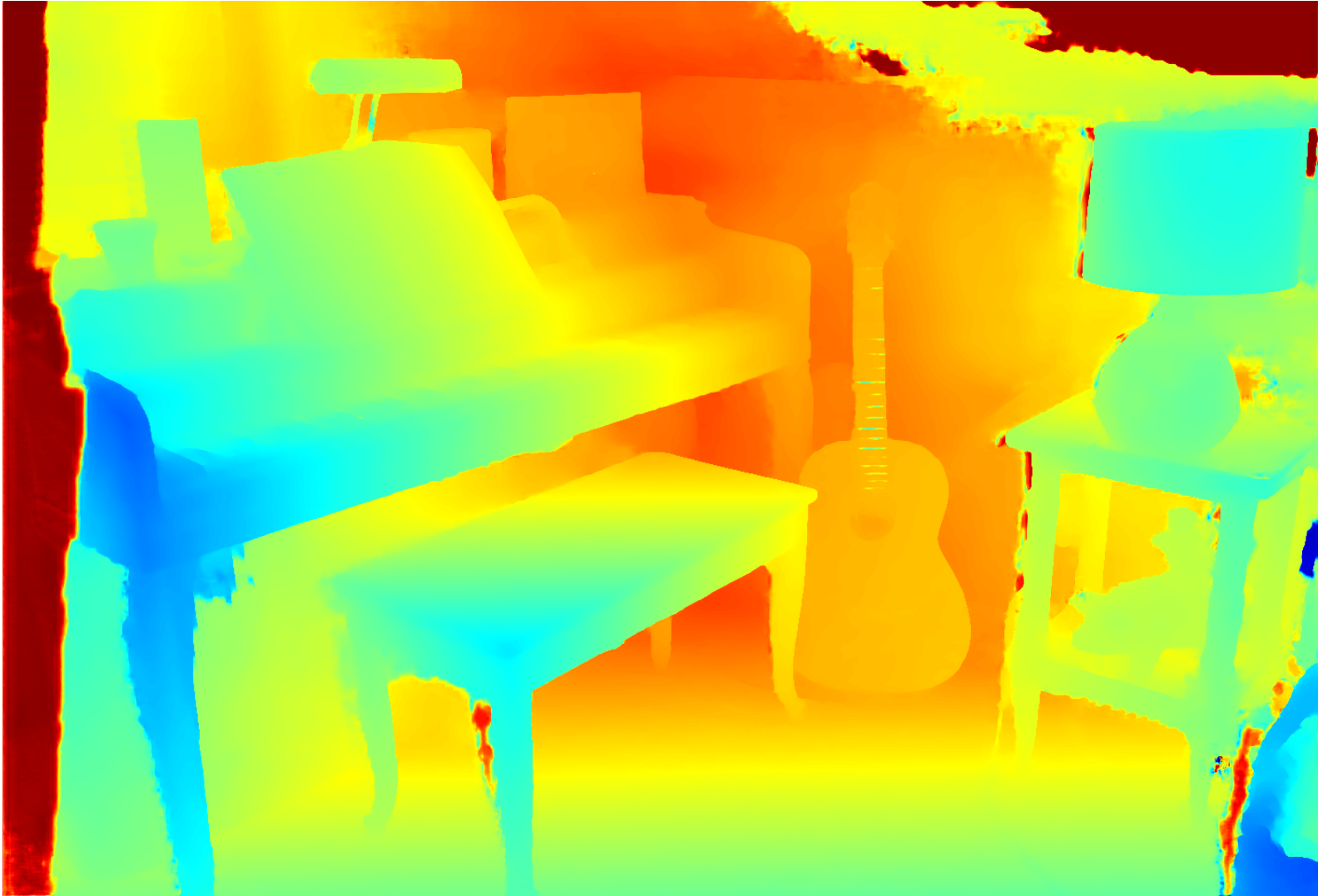}\hspace{0mm}
		\label{chutian7}
          \vspace{-3mm}
	\end{minipage}
	\begin{minipage}{0.241\linewidth}
		\centering
		\includegraphics[width=1.0\linewidth]{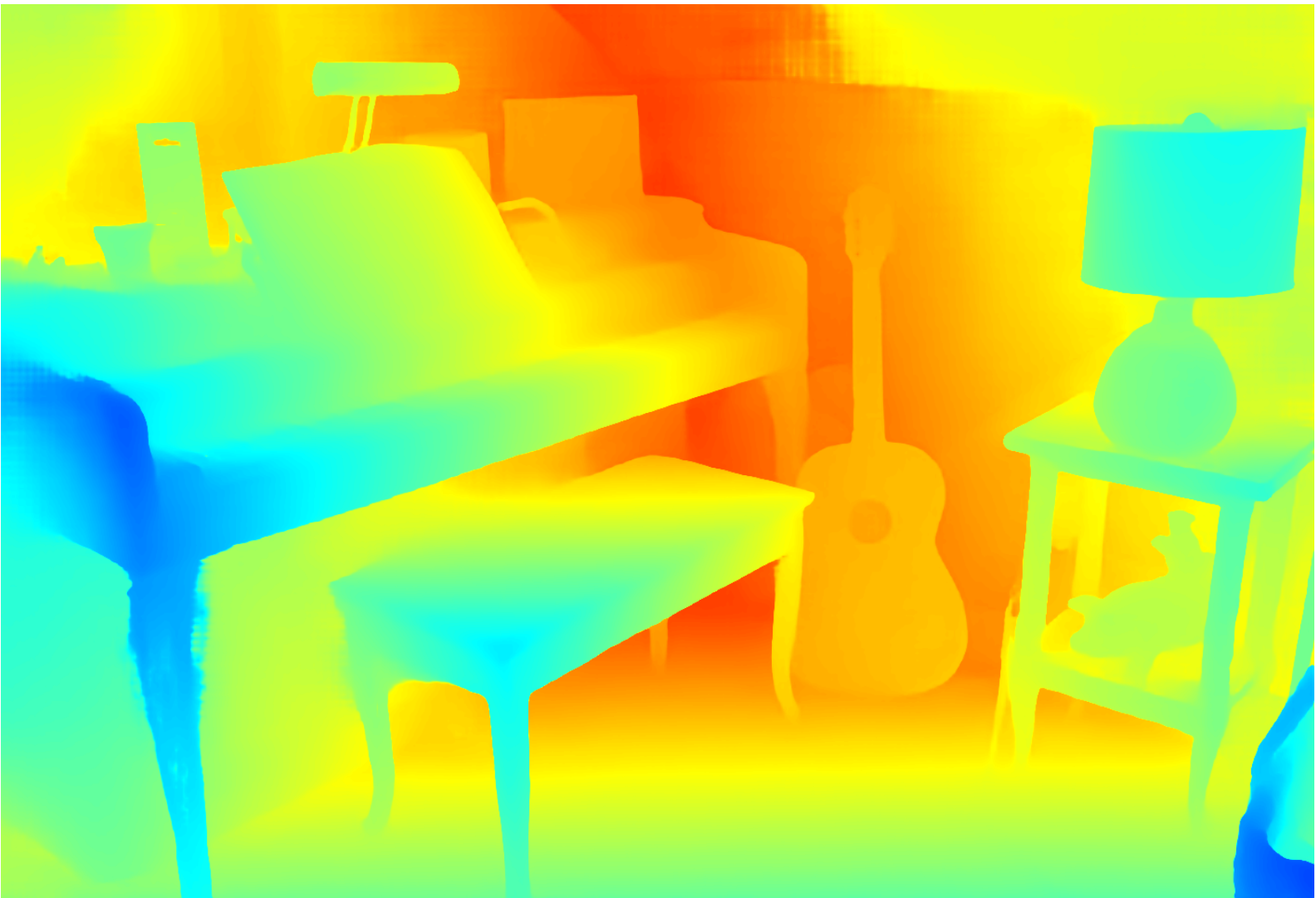}
		\label{chutian8}
                \vspace{-3mm}
	\end{minipage}
	\begin{minipage}{0.241\linewidth}
		\centering
		\includegraphics[width=1.0\linewidth]{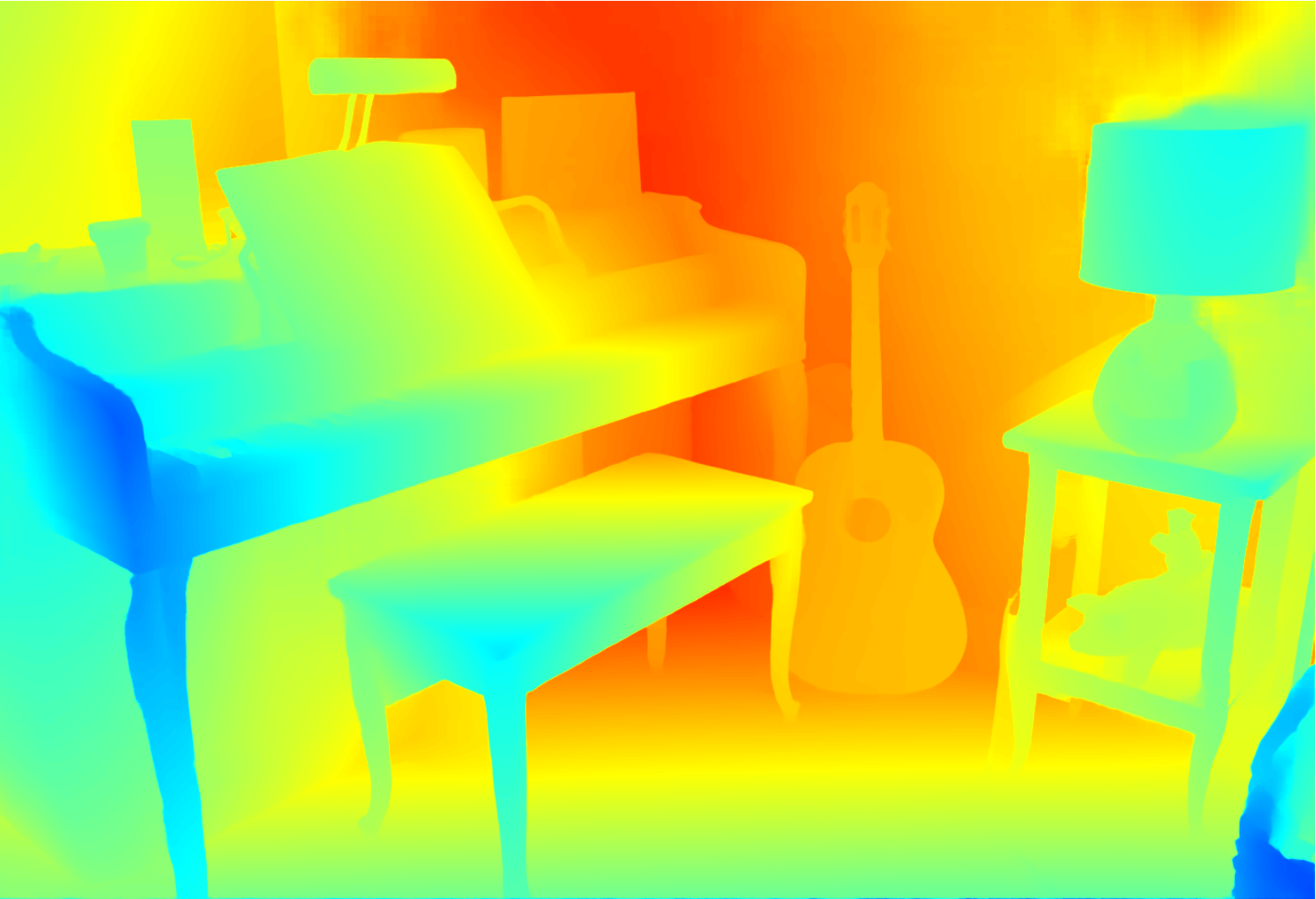}
		\label{chutian9}
                \vspace{-3mm}
	\end{minipage}
\setlength{\abovecaptionskip}{1.mm}
     \caption{Generalization evaluation on Middlebury Dataset.}
    \label{fig:compared_mid}
    \vspace{-4mm}
\end{figure}

It is noteworthy that the KITTI dataset lacks LiDAR ground truth for the upper portions of the images as shown in Figure \ref{fig:compared_with_pcwnet}, s.t. these parts of results are not evaluated in D1-All error. This lack of annotation may introduce bias into the final D1-All error, preventing a complete revelation of the network's effectiveness. Figure \ref{fig:compared_with_pcwnet} illustrates this challenge by comparing the proposed \textit{GOAT} with the most advanced IGEVStereo~\cite{IGEV-Stereo}. Although \textit{GOAT} produces a larger D1-All error, the visualization results exhibit clearly better structures and fewer artifacts in regions where the ground-truth disparity is missing.\\
\begin{table}[!t]
    \setlength{\tabcolsep}{1.3mm}
\caption{Quantitative generalization evaluation on the Middlebury training dataset. \textbf{"Occ"} represents occluded regions, and \textbf{"Non"} represents non-occluded regions. Note the \textbf{\textcolor[rgb]{1,0,0}{Red Bold}} means the best and the \textbf{Bold} means the second-best. }
    \centering
    \scalebox{0.85}{
    \begin{tabular}{c|c|c|c|c|c|c|c|c} \hline
         \multirow{2}{*}{Method} & \multicolumn{2}{c|}{AvgErr} & \multicolumn{2}{c|}{RMSE} & \multicolumn{2}{c|}{Bad 4.0} & \multicolumn{2}{c}{Bad 2.0} \\ \cline{2-9}
         & Occ & Non & Occ & Non & Occ & Non & Occ & Non \\ \hline
         AANet \cite{AANet} & 9.9 & 5.5 & 15.3 & 10.8 & 39.5 & 28.2  & 56.4 & 28.3\\ 
         PSMNet \cite{psmnet} & 17.7 & 10.7 & 29.9 & 22.1  & 47.4 & 23.3 & 62.1 & 32.3\\ 
         GwcNet \cite{gwcnet} & 10.3 & 6.3 & 17.6 & 13.7  & 34.1 & 15.1 & 47.9 & 21.9 \\ 
         ACVNet \cite{wang2020fadnet} & 9.4 & 6.3 & 16.4 & 14.2 & 30.2 & 13.9 & 43.4 & 19.0 \\ 
         PCW-Net \cite{PCW-Net} & \textbf{7.7}& 3.9 & \textbf{14.9} & 9.3  & \textbf{\textcolor[rgb]{1,0,0}{26.5}} & 9.7& \textbf{\textcolor[rgb]{1,0,0}{39.1}}& 14.9\\ 
         STTR-light \cite{STTR} & 35.2& \textbf{3.0} & 47.7 & 10.2 & 74.7 & \textbf{ \textcolor[rgb]{1,0,0}{8.3}} & 82.0 & \textbf{\textcolor[rgb]{1,0,0}{13.3}} \\ 
         RAFTStereo \cite{RaftStereo} & 10.0 & 3.6 & 15.9 & \textbf{9.1}& 34.4 & 9.5 & 46.5 & \textbf{14.4} \\ \hline
         \textbf{GOAT(Ours)} & \textbf{\textcolor[rgb]{1,0,0}{5.7}} & \textbf{\textcolor[rgb]{1,0,0}{2.0}} & \textbf{\textcolor[rgb]{1,0,0}{9.4}} & \textbf{\textcolor[rgb]{1,0,0}{5.3}}  &\textbf{28.0} &\textbf{9.2}& \textbf{43.3} &15.7 \\ \hline
    \end{tabular}
    }
    \label{tab:generalization_mid}
    \vspace{-5mm}
\end{table}
\noindent\textbf{Middlebury.} As the Middlebury dataset only includes 23 images for training, we first evaluate the generalization of the pre-trained SceneFlow model on the Middlebury training set with half resolution. As depicted in Table \ref{tab:generalization_mid}, the proposed \textit{GOAT} generates the best-performing disparity map with the lowest AvgErr and RMSE compared to other methods. Especially in occluded regions, proposed \textit{GOAT} outperforms the latest PCW-Net~\cite{PCW-Net} by 26\% in terms of AvgErr and 33.6\% in terms of RMSE. Figure~\ref{fig:compared_mid} shows the visual comparison. Besides, we also fine-tune our model on the Middlebury dataset with half resolutions (H) because of memory issues. As depicted in Table~\ref{tab:middlebury_fine_tune_half}, compared with other competing methods submitted at the same resolution, our \textit{GOAT} demonstrates state-of-the-art performance by showing the smallest AvgErr and RMSE. Please refer to the \textit{supplementary material} for more results on datasets.
\begin{table}[!ht]
    \centering
    \setlength{\tabcolsep}{1.0mm}
    \caption{Fine-Tuned  Results on Middlebury Benchmarks with half resolution in \textbf{'all'} regions. \textbf{\textcolor[rgb]{1,0,0}{Red Bold: Best}}, \textbf{Bold: Second}.}
\scalebox{0.92}{
    \begin{tabular}{c|c|c|c|c} \hline 
    Method & AvgErr & RMSE & Bad 4.0 & Bad 2.0 \\ \hline
    CFNet~\cite{cfnet} & 5.07&18.20 &11.30 &16.10 \\
    LEAStereo~\cite{leastereo} &\textbf{2.89} &\textbf{13.70} &\textbf{\textcolor[rgb]{1,0,0}{6.33}} &12.10\\
    AANet++~\cite{AANet} &9.77 &24.90&16.40 &22.00 \\
    NOSS\_ROB~\cite{noss_rob}&4.80 & 19.80&8.37 &\textbf{\textcolor[rgb]{1,0,0}{11.20}} \\
    LocalExp~\cite{LocalExp}&5.13 &21.10&8.83 &\textbf{11.30}\\
    FADNet\_RVC~\cite{LocalExp}& 21.00&48.30&24.20 &33.30\\
    MC-CNN-acrt~\cite{MC-CNN}&17.90&55.00&15.80 &19.10 \\
    HITNet~\cite{HINET} &3.29 &14.50&8.66&12.80\\
    ACVNet~\cite{xu2022acvnet} &12.10 &38.60 &12.60 &19.50 \\ \hline
    \textbf{GOAT~(Ours)} &\textbf{\textcolor[rgb]{1,0,0}{2.71}} &\textbf{\textcolor[rgb]{1,0,0}{11.20}}& \textbf{8.18}&13.80 \\ \hline
    \end{tabular}}
    \vspace{-4mm}
    \label{tab:middlebury_fine_tune_half}
\end{table}

\section{Conclusions}
In this paper, we have proposed a novel attention-based stereo-matching network called \textit{GOAT} that exploits long-range dependency and global context for disparity estimation in ill-conditioned regions. The parallel disparity and occlusion estimation module \textit{(PDO)} is proposed to estimate the initial disparity and the occlusion with a parallel attention mechanism, which improves the disparity estimation performance as well as provides the occlusion mask for further disparity refinement. The iterative occlusion-aware global aggregation module (\textit{OGA}) uses a restricted global correlation with a focus scope marked by the occlusion mask to refine the disparity in the occluded regions. Extensive experiments on various datasets have demonstrated the effectiveness and generalization ability of the proposed method. By the time we finish this paper, our method outperforms recent state-of-the-art methods on the SceneFlow dataset and also ranks $1^{st}$ on the KITTI 2015 leaderboard for foreground objects.

{\small
\bibliographystyle{ieee_fullname}
\bibliography{egbib}
}

\end{document}


\title{Global Occlusion-Aware Transformer for Robust Stereo Matching \\
Supplementary Material
}
\author{Zihua Liu$^{1}$, Yizhou Li$^{2}$, and Masatoshi Okutomi$^{3}$ \\
Tokyo Institute of Technology, Japan \\
{\tt\small \{zliu$^{1}$,yli$^{2}$\}@ok.sc.e.titech.ac.jp,mxo@ctrl.titech.ac.jp$^{3}$}
}

\maketitle

\section{More Implementation Details}
\subsection{Details About the Context Adjustment Layer.}
\begin{figure}[!ht]
	\centering
\includegraphics[width=1.0\linewidth]{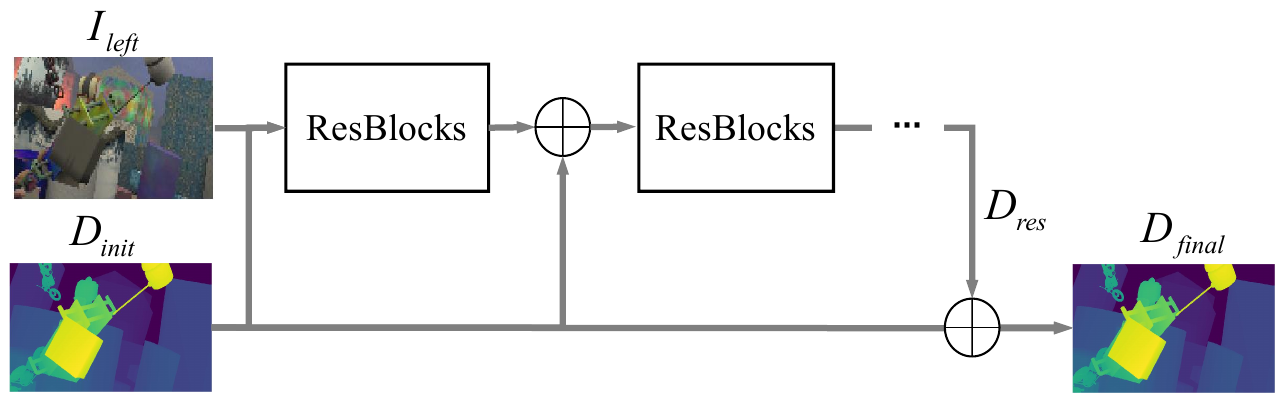}
	\caption{Context Adjustment Layer}
	\label{fig:calayer}
	\vspace{-3 mm}
\end{figure}

The context adjustment layer in \textit{GOAT} is designed to refine the disparity map from a mono-depth aspect. We employ a similar architecture adopted in STTR~\cite{STTR}, which is a simple refinement module comprised of multiple ResBlocks~\cite{resnet}. The architecture of the context adjustment layer is demonstrated in Figure \ref{fig:calayer}.
It can recover disparity details simply by using the left image $I_{left}$ and the current disparity $I_{init}$ as the guidance to regress the disparity residual $D_{res}$ and derive the final disparity $D_{final}$. 
Such an image-based refinement module can help refine the disparities in extremely large occluded regions where no matching clues can be utilized.
\subsection{Occlusion Mask Generation for Supervision}
\paragraph{SceneFlow:} Because the SceneFlow \cite{dispnetc} dataset's ground-truth disparity for the left view and right view are both annotated for all pixels, it becomes feasible to generate dense ground-truth occlusion mask $M_{occ}$ by directly applying left-right consistency check~\cite{okutomi_kanede}. The process can be described as follows:
\begin{equation}
M_{occ}=
\begin{cases}
1& \text{if $D_{gap}\geq1,$}\\
0& \text{otherwise,}
\end{cases} 
\end{equation}
\begin{equation}
D_{gap} = \lvert D_{L}(x,y) - D_{R}(x+D_{L}(x,y),y) \rvert ,
\end{equation}
where $D_{gap}$ is the disparity difference between the corresponding pixels at the left and right views, and $D_L$ and $D_R$ are ground-truth disparity maps for left and right views, respectively.
\begin{figure}[!t]
	\centering
\includegraphics[width=1.0\linewidth]{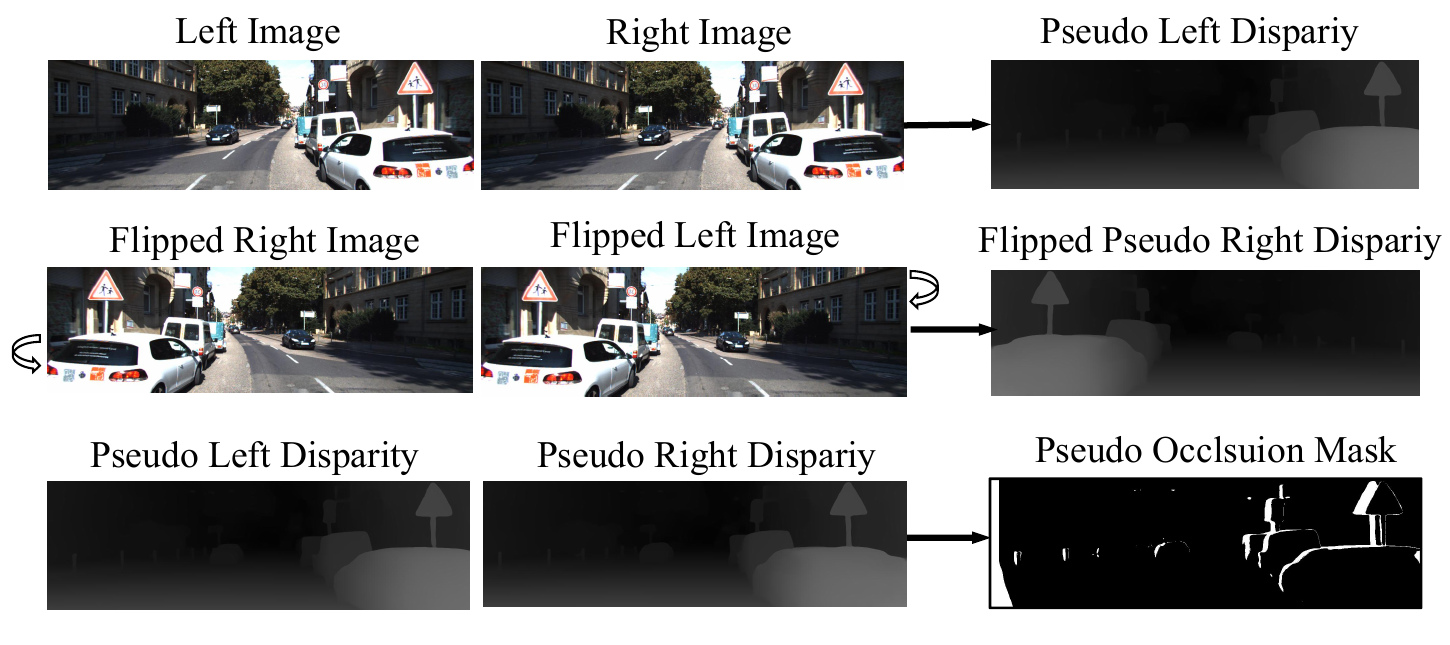}
\caption{Flipped inference consistency check for pseudo occlusion mask generation.}
	\label{fig:flipped_inference}
	\vspace{-5 mm}
\end{figure}
\begin{figure}[!ht]
	\centering
\includegraphics[width=1.0\linewidth]{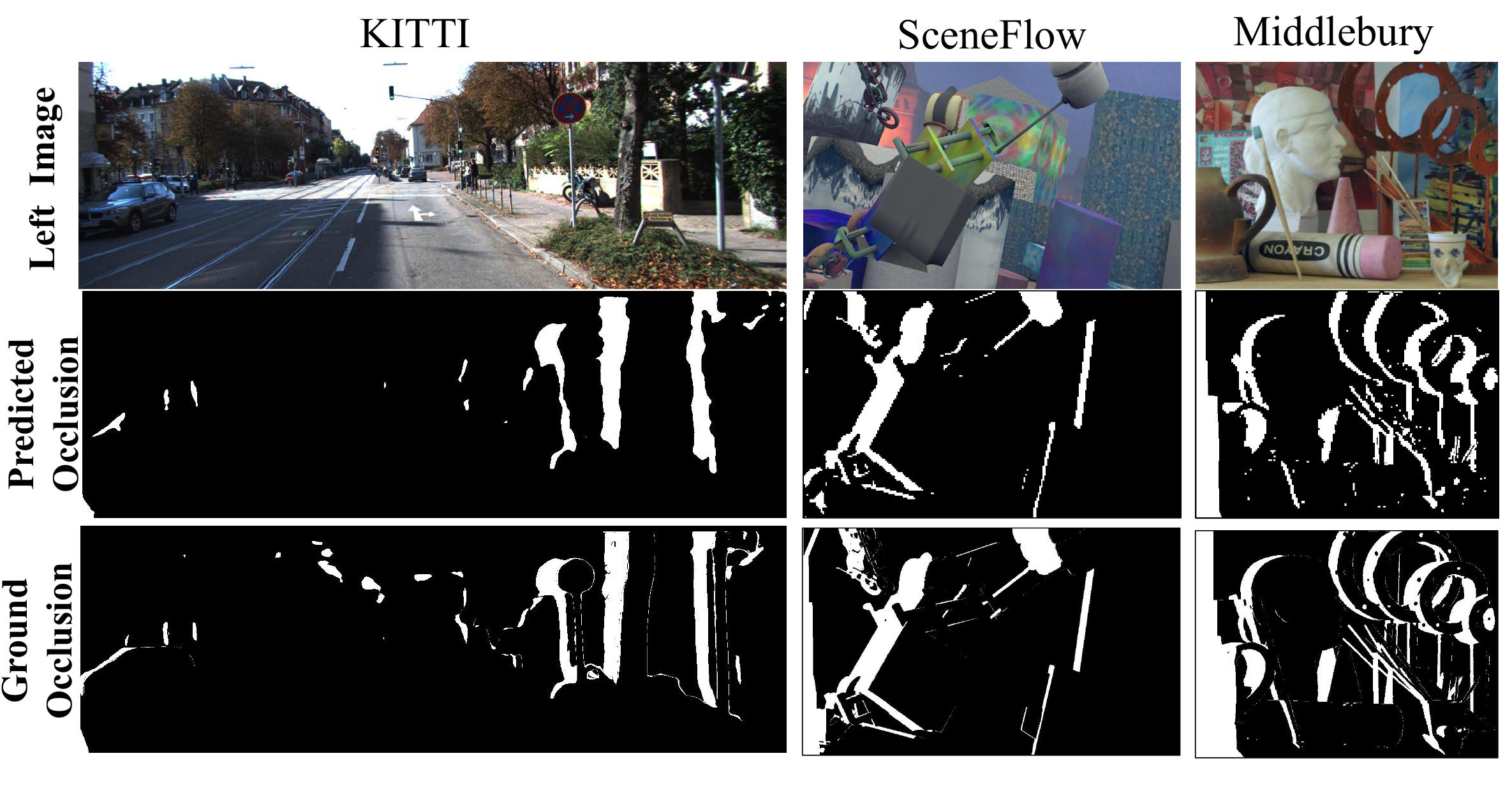}
\caption{Predicted occlusion masks and corresponding ground truth on different datasets.}
	\label{fig:occlusion_estimation_results}
	\vspace{-5 mm}
\end{figure}
\paragraph{KITTI and Middlebury:}
Generating the ground-truth occlusion mask for the KITTI dataset~\cite{Kitti2015} poses two significant challenges: Firstly, KITTI's ground-truth disparity maps are derived through LiDAR, resulting in sparse disparity annotation. Secondly, the KITTI dataset only provides the ground-truth disparity maps for the left images, which makes it difficult to directly apply the left-right consistency check to generate the occlusion masks. The STTR~\cite{STTR} attempts to mitigate these challenges by solely focusing on the out-of-bound occlusion mask, which comprises only the pixels that fall outside the field of view~(FOV) of the right image. However, this method does not provide sufficient supervision for the occlusion mask estimation. To tackle this problem, we design a pipeline named flipped inference consistency check to generate the pseudo occlusion masks.

As demonstrated in Figure \ref{fig:flipped_inference}, we leverage a model pre-trained on the KITTI dataset without occlusion supervision as the disparity generator. Initially, the left-right image pair is employed as input to produce a dense pseudo-left disparity. Subsequently, the left view image is horizontally flipped to create a new right view, and similarly, the right view image is flipped to generate a new left view. The new left-right image pair will be sent to the identical disparity generator, where the flipped pseudo-right disparity is obtained. Finally, the left-right consistency check between the pseudo disparity maps for the left view and the right view is applied to generate a pseudo occlusion mask.
For Middlebury dataset~\cite{middlebury2014}, we use the same strategy for occlusion mask generation. Figure~\ref{fig:occlusion_estimation_results} shows some examples of estimated occlusion masks by proposed \textit{GOAT} and their corresponding ground truth on the SceneFlow, KITTI, and Middlebury datasets.
\section{More Training Details}
In this section, we present further information regarding the training process on different datasets.
\subsection{Data Augmentation}
A domain gap exists between synthetic SceneFlow datasets and real-world KITTI and Middlebury datasets in terms of color and disparity distribution. This poses a challenge for fine-tuning with SceneFlow pre-trained models, which is further compounded by limited annotated training data in real-world KITTI and Middlebury datasets. To enhance the network's robustness and mitigate overfitting, we employ augmentations as follows. \\
\textbf{Asymmetric and Symmetric Chromatic Augmentations:} To address the issue of diverse lighting and exposure conditions in real-world stereo images, we adopted a method similar to that used in HSM~\cite{HSM}. This involved modifying the brightness, contrast, and gamma of both left and right images with random adjustment parameters from intervals of $[0.8,1.2]$, with the option of using different parameters for the left and right images. This allowed us to simulate color and exposure variations commonly observed in real scenes.\\
\textbf{Color Domain Adaption: }\label{cda} To alleviate the difference in color distribution between synthetic data and real-world data, we used color domain adaptation augmentation following \cite{adastereo}. This method utilizes normalization techniques in the LAB color space to reduce the distribution gap between the two types of data.\\
\textbf{Vertical y-offset and Flip:}
To simulate the disparity drift problem caused by imperfect calibration, we applied the y-offset augmentation from \cite{HSM}, which randomly shifts the y-direction pixels in the right image by an offset within [-2,2] pixels. We also utilized symmetric vertical flipping for both left and right views to improve disparity estimation accuracy across all image regions and prevent location bias.\\
\textbf{Asymmetric Masking: }
Similar to \cite{HSM}, we replaced the random patches of the right images with mean values of the whole images. By applying this, it will increase the proportion of occluded areas, making the Occlusion-Aware Global Aggregation Module~(\textit{\textit{OGA}}) more effective. The size of the patch to be replaced was randomly sampled between [40,40] and [120,180].
\subsection{Training Setup}\label{trainingsetup}
\noindent \textbf{SceneFlow:} As described in the paper, for training on the SceneFlow dataset, we use all three sub-sets~(Flyingthings3D, Driving, and Monkaa) within a total of 35K images. We consider a random crop of $320\times640$ with a batch size of 8 and a maximum disparity of 192. The whole training process on SceneFlow is performed with 4 NVIDIA RTX 3090 GPUs without data augmentation. \\
\textbf{KITTI:}
In regards to the fine-tuning on the KITTI 2015 dataset, the color domain adaption described in Section \ref{cda} was initially employed to fine-tune the pre-trained model for an additional 40 epochs on the SceneFlow dataset, utilizing a learning rate of 1e-4. Following this, we used mixed datasets containing KITTI 2012 and KITTI 2015 with in total of 400 pairs of images for the training of the first 400 epochs. We chose the model with the best performance on the validation set, followed by another 200 epochs of fine-tuning on the KITTI 2015 training set to obtain the final model. The whole training process was conducted on 2 NVIDIA RTX6000 GPUs, with a patch size of 320$\times$1088. \\
\textbf{Middlebury: }To address the limited amount of training data in the Middlebury dataset, we followed the similar strategy utilized in \cite{GREStereo} by augmenting the Middlebury dataset to the 20$\%$ amount of the SceneFlow dataset with techniques mentioned in \ref{cda}. We conducted mixed training for 100 epochs and then fine-tuned the model on the Middlebury dataset alone for another 500 epochs to get the final model.
\begin{figure*}[!t]
    \centering  \includegraphics[width=0.9\linewidth]{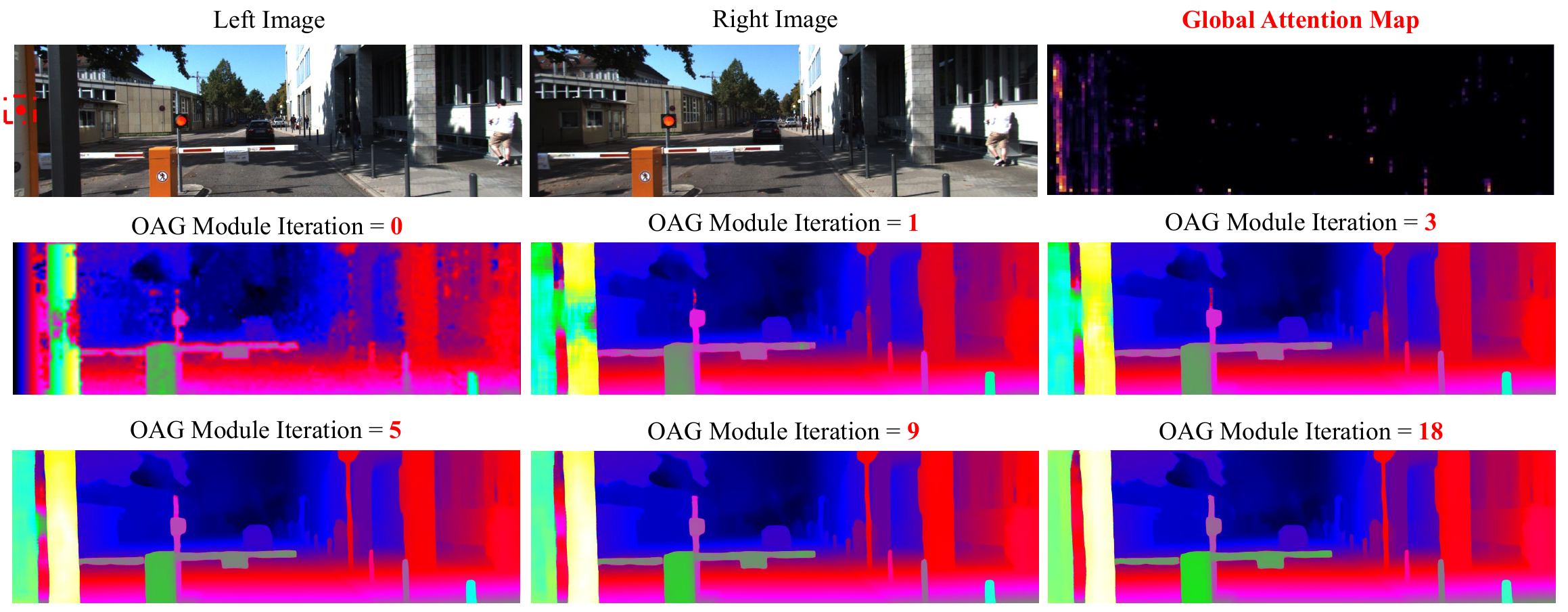}
    \caption{ \textbf{Intermediate Outputs for \textit{OGA} module.} The third image of the first row illustrates the global attention map of the location marked by a red dot in the left image. Images in the second row and the third row demonstrate the disparity estimation results at the different iterations using the proposed \textit{OGA} module. Through iterative refinement using the \textit{OGA} module, the disparity estimation improves progressively, particularly in occluded regions.}
    \label{fig:intermed}
\end{figure*}
\section{Intermediate Outputs for the OGA Module}
  Similar to \cite{RaftStereo}, the proposed \textit{OGA} module is a GRU-based iterative refinement module. To further demonstrate how the \textit{OGA} module uses global correlation to refine the disparities in the occluded area, we illustrate intermediate disparity outputs of the \textit{OGA} module of each iteration in the KITTI dataset in Figure \ref{fig:intermed}. 
  
  The \textit{OGA} module applies a global attention mechanism to aggregate features within occluded regions, thereby enhancing the accuracy of disparity estimations. Our findings indicate that as the number of iterations of the \textit{OGA} module increases, the estimated disparity progressively improves. It is worth noting that applying the \textit{OGA} module only once can already yield considerable enhancement in the disparity estimation compared with the preliminary results as shown in Figure \ref{fig:intermed}.

\section{Additional Experiments Results}
\subsection{More Comparisons on the SceneFlow dataset}
`In this subsection, we will provide more comparisons with the proposed \textit{GOAT} and other SOTA methods on the SceneFlow dataset. Based on the results presented in Table \ref{tab:supp_sf}, it is apparent that the proposed \textit{GOAT} surpasses all other evaluated approaches in regards to both the overall end-point-error~(EPE) and the EPE at the occluded regions(EPE-Occ). Furthermore, the proposed approach demonstrates a lower rate of outliers, as indicated by P1 and P3 values.

\begin{table}[!t]
    \centering
    \caption{we compare the performance of our proposed methods with other notable works on the \textbf{SceneFlow} dataset. We present the end-point-error (EPE) results for disparities in overall (All) regions, as well as occluded regions (Occ), and report P1 and P3 errors. The best-performing method is highlighted in \textbf{Bold}.}
     \scalebox{0.9}{
    \begin{tabular}{c|cc|c|c} \hline
    \multirow{2}{*}{\quad Method   } & \multicolumn{2}{c|}{EPE}  &\multirow{2}{*}{P1 Error}  &\multirow{2}{*}{P3 Error} \\ 
    \cline{2-3} & EPE-All & EPE-Occ & &\\ \hline
    DispNetC~\cite{dispnetc} & 1.68 &- & -& - \\
    StereoNet~\cite{stereonet} & 1.07 &3.31 &13.7$\%$ & 5.3$\%$ \\
    AANet++~\cite{AANET}  & 0.72 &2.44 & 10.4$\%$ & 4.0$\%$\\
    PSMNet~\cite{psmnet}& 1.09 &3.14 & 11.1$\%$ & 4.6$\%$ \\
    GANet~\cite{ganet}& 0.84 &2.83 & 8.8$\%$ & 3.8$\%$\\
    GwcNet~\cite{gwcnet}& 0.77 &2.47 & 8.7$\%$& 3.9$\%$ \\
    RAFT-Stereo~\cite{RaftStereo} &0.69 &2.14 & 7.9$\%$& 3.3$\%$ \\
    STTR-light~\cite{STTR} &4.14 &23.9 &16.4$\%$ & 10.7$\%$ \\
    ACVNet~\cite{xu2022acvnet} &0.48&1.65& \textbf{6.2$\%$}& 2.9$\%$\\
    EDNet~\cite{zhang2021ednet} &0.63 &2.08 & 8.2$\%$& 3.9$\%$ \\
    IGEVStereo~\cite{xu2022acvnet} &\textbf{\textcolor[rgb]{1,0,0}{0.47}}&1.62& 6.6$\%$& 3.3$\%$\\
    PCW-Net~\cite{PCW-Net} &0.86 &2.54 & 9.1$\%$& 4.0$\%$ \\
    \hline
    \textbf{GOAT~(Ours)}&\textbf{\textcolor[rgb]{1,0,0}{0.47}} &\textbf{\textcolor[rgb]{1,0,0}{1.53}}& \textbf{\textcolor[rgb]{1,0,0}{5.6$\%$}} & \textbf{\textcolor[rgb]{1,0,0}{2.7$\%$}}\\
    \hline
    \end{tabular}
    }
    \label{tab:supp_sf}
    \vspace{-1mm}
\end{table}
\begin{table}[!t]
    \centering
    \caption{Numbers of training parameters (Params) and multi-accumulate operations~(Macs) compared with other lastest methods. We use an input resolution of 320$\times$640 for Mac's computation.}
     \scalebox{1.0}{
    \begin{tabular}{c|c|c|c} \hline
     Method 
    & Params 
    & Macs 
    & EPE \\ \hline
     \quad ACVNet~\cite{xu2022acvnet} \quad  &  7.1M & 465.1G  & 0.48  \\ 
     \quad RAFTStereo~\cite{RaftStereo} \quad &  11.1M & 654.8G  & 0.69  \\ 
     \quad PCW-Net~\cite{PCW-Net} \quad &  35.8M & 768.6G  & 0.86  \\ 
     \quad IGEVStereo~\cite{IGEV-Stereo} \quad &  12.6M & 541.6G  & 0.47  \\ 
     \hline
     \textbf{GOAT-T~(Ours)} &  10.0M & 192.0G  & 0.56  \\ 
     \textbf{GOAT~B(Ours)} &  12.1M & 858.3G  & 0.47  \\ 
    \hline
    \end{tabular}
    }
    \vspace{-2mm}
    \label{tab:mac_para}
\end{table}

\begin{table*}[!tp]
\centering
\caption{Ablation study of our proposed \textit{GOAT} on the FallingThings~\cite{Fallingthings3D} Dataset. \textit{"PDO"} is short for Parallel Disparity and Occlusion Estimation Module. \textit{"OGA"} is short for Iterative Occlusion-Awareness Global Aggregation Module. We calculated the EPE and P1(outliers) both in the overall and the occluded regions, separately."*" means a higher resolution.} 
\scalebox{0.92}{
\begin{tabular}{cl|cc|cc|c|c|cc|cc|c|c}
\hline
\multicolumn{2}{c|}{\multirow{2}{*}{Method}} &
  \multicolumn{2}{c|}{\begin{tabular}[c]{@{}c@{}}Disparity \\ Estimation\end{tabular}} &
  \multicolumn{2}{c|}{\begin{tabular}[c]{@{}c@{}}Update\\ Module\end{tabular}} &
  \multirow{2}{*}{\begin{tabular}[c]{@{}c@{}}CA\\ Layer\end{tabular}} &
   &
  \multicolumn{2}{c|}{EPE} &
  \multicolumn{2}{c|}{P1(\%)} &
  \multirow{2}{*}{\begin{tabular}[c]{@{}c@{}}Occ\\ mIOU\end{tabular}} &
  \multirow{2}{*}{Res} \\ \cline{3-6} \cline{9-12}
\multicolumn{2}{c|}{} &
  \begin{tabular}[c]{@{}c@{}}Cost\\ Volume\end{tabular} &
  PDO &
  RAFT &
  OGA &
   &
   &
  All &
  Occ &
  All &
  Occ &
   &
   \\ \cline{1-7} \cline{9-14} 
\multicolumn{2}{c|}{Baseline} &
  \checkmark &
   &
  \checkmark &
   &
   &
   &
  0.53 &
  2.30 &
  5.9\% &
  28.9\% &
  - &
  1/8 \\ \cline{1-7} \cline{9-14} 
\multicolumn{2}{c|}{PDO} &
   &
  \checkmark &
  \checkmark &
   &
   &
   &
  0.41 &
  1.65 &
  5.2\% &
  26.7\% &
  0.905 &
  1/8 \\
\multicolumn{2}{c|}{PDO~+~OGA} &
   &
  \checkmark &
   &
  \checkmark &
   &
   &
  0.31 &
  1.22 &
  3.6\% &
  20.2\% &
  0.905 &
  1/8 \\
\multicolumn{2}{c|}{PDO~+~OGA~+~CA} &
   &
  \checkmark &
   &
  \checkmark &
  \checkmark &
   &
  0.29 &
  1.18 &
  3.4\% &
  19.1\% &
  0.906 &
  1/8 \\
\multicolumn{2}{c|}{PDO~+~OGA~+~CA*} &
   &
  \checkmark &
   &
  \checkmark &
  \checkmark &
   &
  0.25 &
  1.11 &
  2.7\% &
  17.6\% &
  0.913 &
  1/4 \\ \hline
\end{tabular}}
\label{tab:AblationStudyFallingThings}
\end{table*}
\begin{figure*}[!h]
    \centering
    \vspace{-3mm}
\includegraphics[width=1.0\linewidth]{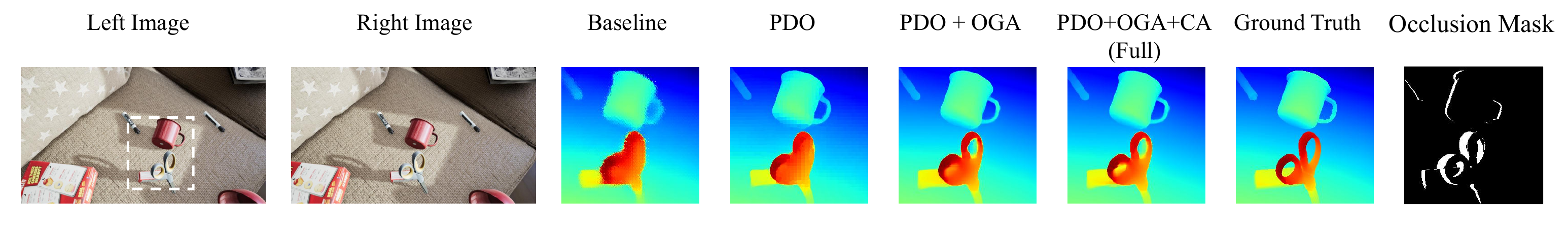}
    \caption{Visualizations of ablation studies on FallingThings Datset. We cropped and enlarge the selected part of the disparity map for easier viewing.}
\label{fig:ablation_fallingthings_vis}
\end{figure*}
Furthermore, an important aspect of efficient neural network architecture is the training parameters (Params) as well as their computing complexities, which are often quantified by the number of multi-accumulate operations (Macs). As shown in Table \ref{tab:mac_para}, our \textit{GOAT-T} with 1/8 resolution can achieve a competitive performance with the small-est Macs and rather small parameters compared with the latest PCWNet~\cite{PCW-Net} and IGEVStereo~\cite{IGEV-Stereo}. However, transformer-based methods inherently suffer from quadratic computational complexity where the \textit{GOAT-B} has 4 times larger Macs when increasing the resolution to 1/4.

\begin{figure*}[!t]
    \centering
\includegraphics[width=1.0\linewidth]{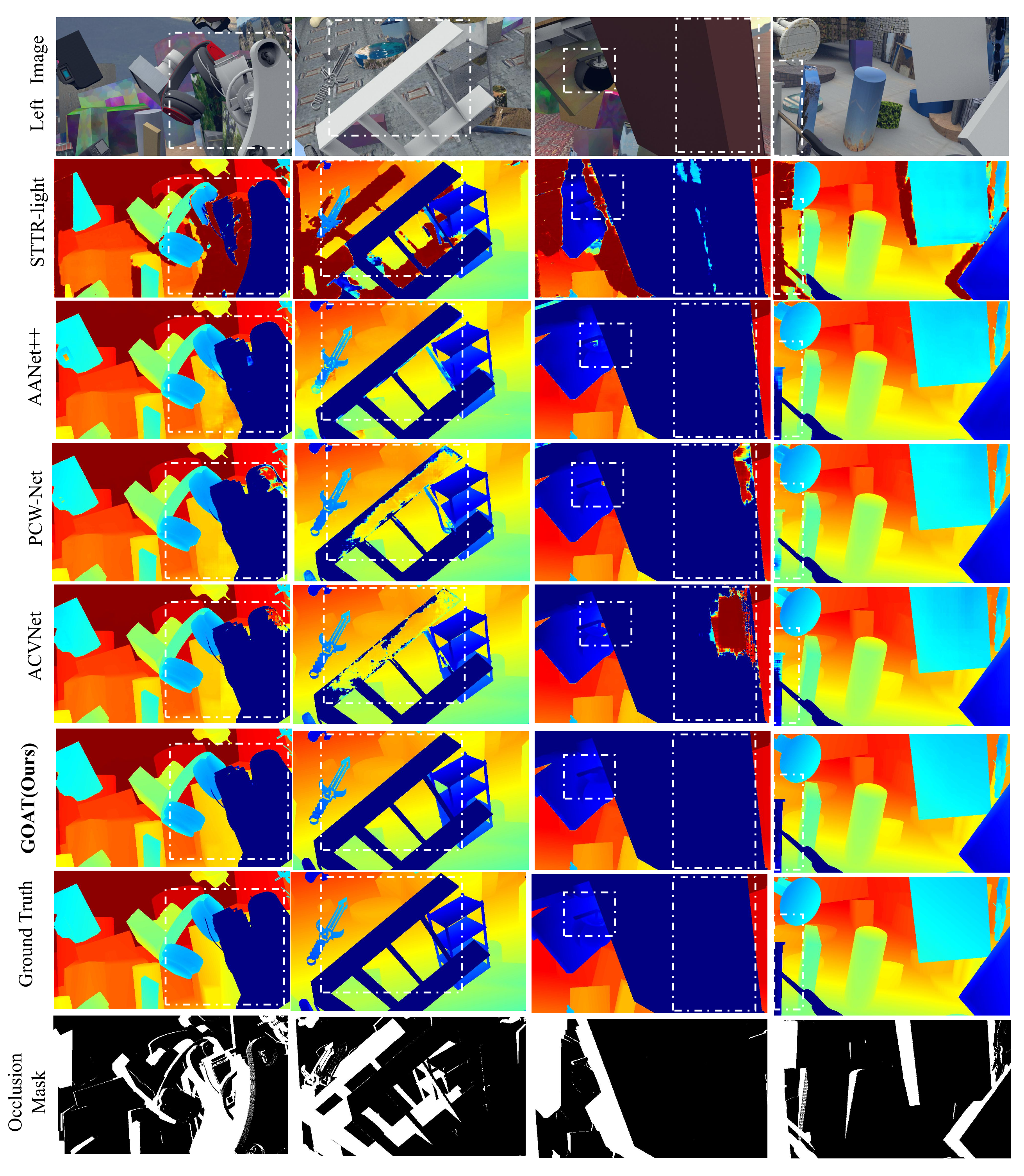}
    \caption{Visualization comparison of estimated disparities on the SceneFlow dataset. Our proposed \textit{GOAT} demonstrates more structured and continuous disparity results in the white bounding box. } \label{fig:sf_comparsion_supp}
\end{figure*}

We present a comprehensive analysis with visualization of the estimated disparity on the SceneFlow dataset which can be inferred in Figure \ref{fig:sf_comparsion_supp}. Compared with other notable networks, the proposed \textit{GOAT} can generate better disparity estimation results around the thin structures and in the texture-less regions with the assistance of the proposed parallel disparity and occlusion estimation module (\textit{PDO}). Besides, our proposed method displays strong robustness in the occluded regions where other notable approaches fail to yield a satisfactory result in such ill-conditioned regions.

\subsection{Supplementary Ablation Studies on the FallingThings Dataset.}
Besides ablation studies shown in the main paper, we also conduct the ablation studies on the FallingThings~\cite{Fallingthings3D} dataset. Compared to random floating objects in SceneFlow dataset, FallingThings dataset contains scenes with carefully placed objects, thus it has more realistic semantics and occlusions. The related results can be shown in Table \ref{tab:AblationStudyFallingThings}.  

Compared with the Baseline, the model integrates with the \textit{PDO} module (designated as PDO) is able to improve the overall performance by a big margin from 0.53 to 0.41. As demonstrated in Figure ~\ref{fig:ablation_fallingthings_vis}, applying the \textit{PDO} shows better structural disparity performance where the baseline shows blur and ambiguous disparity values.  

Furthermore, Table \ref{tab:AblationStudyFallingThings} exemplifies the efficacy of the \textit{OGA} module. It demonstrates an enhancement in the performance of disparity estimation within occluded regions, reducing the disparity from 1.65 to 1.22, resulting in a 26\% improvement.
This pattern is also observable in Figure \ref{fig:ablation_fallingthings_vis}, where the PDO~+~OGA clearly shows better disparities in the occluded regions.

Finally, the complete model incorporated with \textit{PDO} and \textit{OGA} modules witnesses the best performance by showing an EPE-Occ of 1.18.
\subsection{More Comparisons on the KITTI dataset.}
\begin{figure*}[!t]
    \centering
    \includegraphics[width=1.0\linewidth]{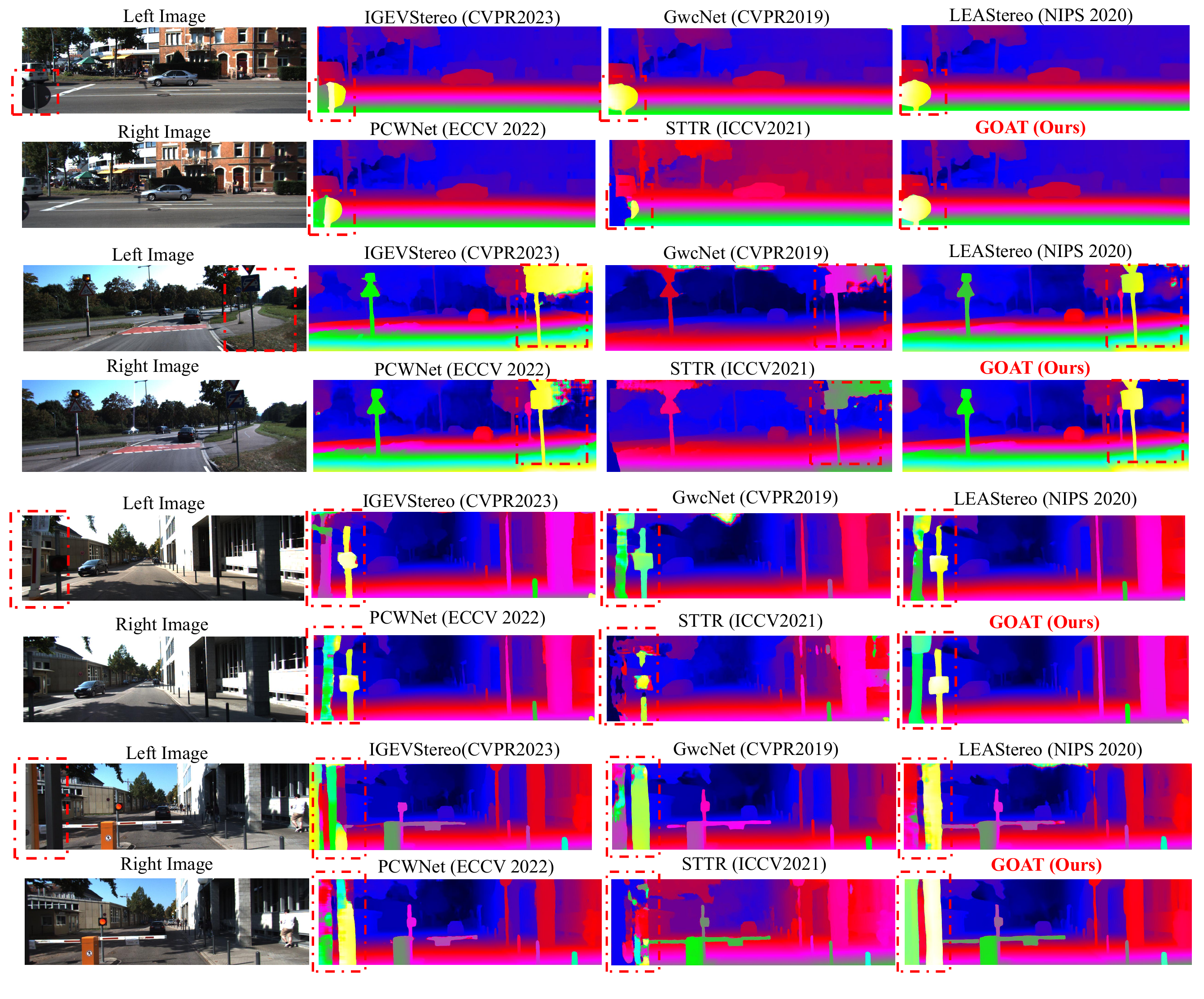}
    \caption{Visualization comparison of estimated disparities on the KITTI 2015 dataset. Note our proposed \textit{GOAT} can generate more detailed disparities outputs, especially in the occluded regions compared with other SOTA networks. }
    \label{fig:kitti_comparsion_supp}
\end{figure*}

In this section, we will provide more visualization comparison results on the KITTI 2015 test set. As illustrated in Figure \ref{fig:kitti_comparsion_supp}, the proposed \textit{GOAT} has more continuous disparity estimation results in the occluded areas (regions within the red bounding boxes.). But other most advanced methods, such as PCWNet~\cite{PCW-Net} and IGEVStereo~\cite{IGEV-Stereo}, have obvious outliers and disparity discontinuities. Furthermore, our proposed \textit{GOAT} model demonstrates superior robustness compared to alternative networks, as evidenced by its ability to produce more accurate structural modeling of street scenes with fewer artifacts. This characteristic is highly advantageous for autonomous driving applications.

\subsection{More Comparisons on the Middlebury dataset.}
\begin{figure*}[!t]
    \centering
    \includegraphics[width=1.0\linewidth]{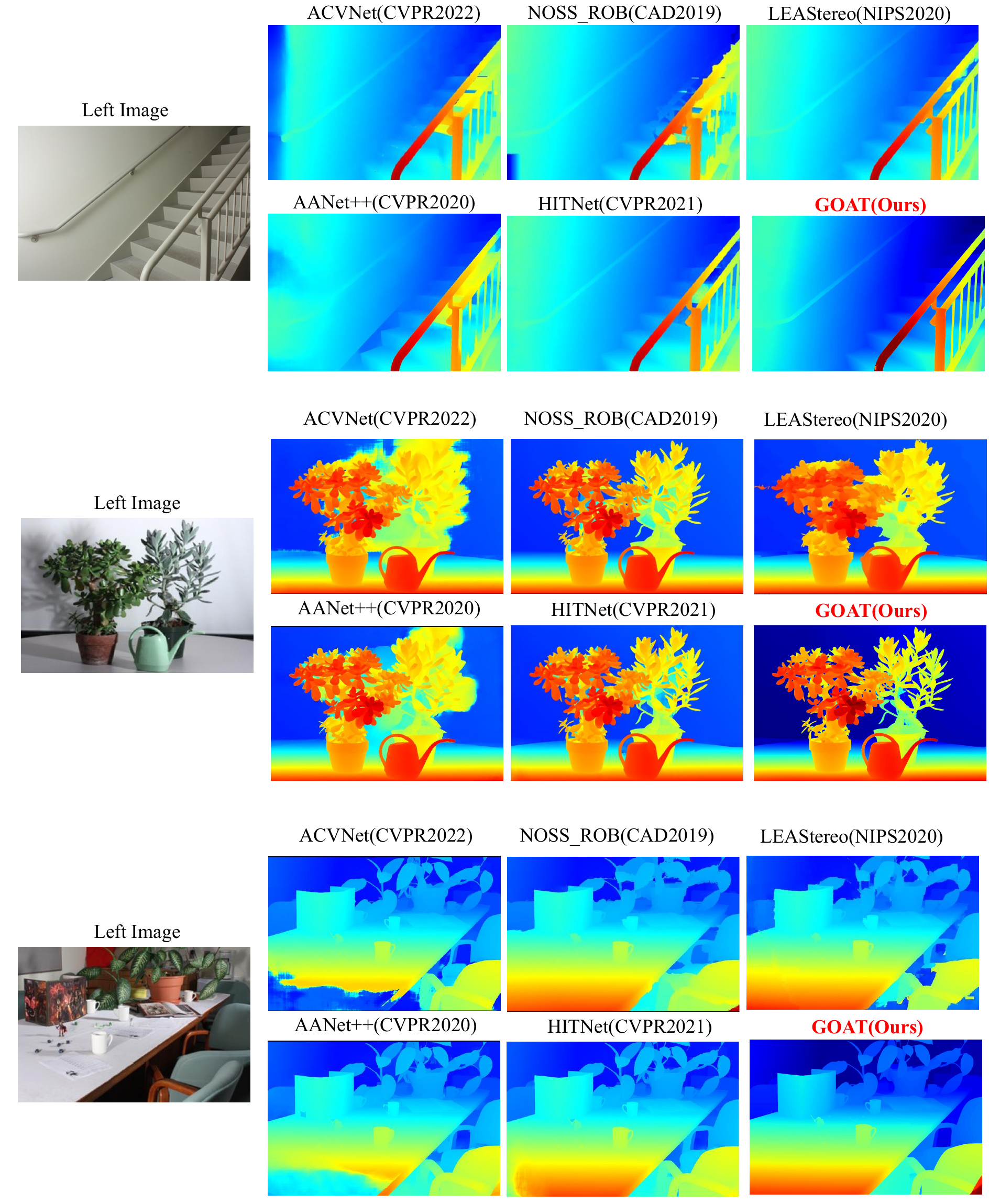}
    \caption{Visualization comparison of estimated disparities on the Middlebury test set. }
    \label{fig:mid_comparsion_supp}
\end{figure*}

In addition to the generalization evaluation visualization mentioned in our paper, we also fine-tuned the SceneFlow pre-trained model on the Middlebury dataset with half resolution (H) following the training scheme outlined in \ref{trainingsetup}. 
More visualization results can be inferred from Figure \ref{fig:mid_comparsion_supp}, where our proposed method demonstrates better structural disparity estimation and fewer artifacts.
\\
\\

{\small
\bibliographystyle{ieee_fullname}
\bibliography{egbib}
}

